\documentclass[12pt]{article}
\usepackage{amsmath}
\usepackage{times}
\usepackage{graphicx}
\usepackage{color}
\usepackage{multirow}

\usepackage{sectsty}
\sectionfont{\fontsize{12}{15}\selectfont}
\subsectionfont{\fontsize{12}{15}\selectfont}
\subsubsectionfont{\fontsize{12}{15}\selectfont}

\usepackage[authoryear]{natbib}
\usepackage{fancyhdr}
\usepackage{rotating}
\usepackage{bbm}
\usepackage{latexsym}

\usepackage{bm}
\usepackage{wrapfig}
\usepackage{booktabs}
\usepackage{url}       

\newcommand{\captionfonts}{\normalsize}

\makeatletter  
\long\def\@makecaption#1#2{%
  \vskip\abovecaptionskip
  \sbox\@tempboxa{{\captionfonts #1: #2}}%
  \ifdim \wd\@tempboxa >\hsize
    {\captionfonts #1: #2\par}
  \else
    \hbox to\hsize{\hfil\box\@tempboxa\hfil}%
  \fi
  \vskip\belowcaptionskip}
\makeatother   

\begin{document}
\hspace{13.9cm}1

\ \vspace{20mm}\\

{ \LARGE Predictive Coding for Dynamic Visual Processing: Development of Functional Hierarchy in a Multiple Spatio-Temporal Scales RNN Model\\}
\ \\
{\bf Author$^{\displaystyle 1, \displaystyle 2}$}\\
{$^{\displaystyle 1}$\textbf{Minkyu Choi}}\\
School of Electrical Engineering, Korea Advanced Institute of Science and Technology (KAIST), Daejeon, 305-701, Republic of Korea\\
minkyu.choi8904@gmail.com\\
{$^{\displaystyle 2}$\textbf{Jun Tani}}\\
Okinawa Institute of Science and Technology (OIST), Okinawa, Japan 904-0495\\ and School of Electrical Engineering, Korea Advanced Institute of Science and Technology (KAIST), Daejeon, 305-701, Republic of Korea\\
tani1216jp@gmail.com\footnote{The correspondence should be sent to Jun Tani}\\
%


\thispagestyle{empty}
\markboth{}{NC instructions}
\ \vspace{-0mm}\\
%
\begin{center} {\bf Abstract} \end{center}
The current paper proposes a novel predictive coding type neural network model, the predictive multiple spatio-temporal scales recurrent neural network (P-MSTRNN). The P-MSTRNN learns to predict visually perceived human whole-body cyclic movement patterns by exploiting multiscale spatio-temporal constraints imposed on network dynamics by using differently sized receptive fields as well as different time constant values for each layer. After learning, the network becomes able to proactively imitate target movement patterns by inferring or recognizing corresponding intentions by means of the regression of prediction error. Results show that the network can develop a functional hierarchy by developing a different type of dynamic structure at each layer. The paper examines how model performance during pattern generation as well as predictive imitation varies depending on the stage of learning. The number of limit cycle attractors corresponding to target movement patterns increases as learning proceeds. 
And, transient dynamics developing early in the learning process successfully perform pattern generation and predictive imitation tasks. The paper concludes that exploitation of transient dynamics facilitates successful task performance during early learning periods.

\section{Introduction}
Predictive coding is a plausible model to account for how brains can predict future perception. The central notion is that current top-down ‘intention’ as represented in higher level processes can be modified using error between the prior predicted and actually perceived outcome of prior intentional action in the bottom-up pathway \citep{Rao1999, Tani1999, Friston2005, Clark2015}. Within this predictive coding framework, it has been largely assumed that the necessary functional hierarchy develops across multiple cortical regions spanning higher level regions representing intention and lower level regions representing direct perception \citep{Rao1999, Rao2003, Friston2005, Ouden2012}. The current study proposes a novel recurrent neural network (RNN) model in order to examine how a spatio-temporal hierarchy adequate for the robust generation and recognition of dynamically composed visual patterns can be developed within the predictive coding framework. 

Inspired by upward and downward causation \citep{Campbell1974, Bassett2011}, the current study assumes that a spatio-temporal hierarchy is indispensable for the compositional manipulation of visual streams, and that this hierarchy can develop naturally if adequate spatio-temporal constraints are imposed on neural activity, as in the proposed model. The currently proposed model is an extension of a prior model, the multiple spatio-temporal scales RNN (MSTRNN) \citep{Lee2016} used for the categorization of dynamic visual patterns. The MSTRNN, like this newly proposed predictive-MSTRNN (P-MSTRNN), integrates two different ideas, one from the multiple timescale RNN (MTRNN) \citep{Yamashita2008} and the other from convolutional or deconvolutional neural networks (CNN, deCNN) \citep{LeCun1998, Zeiler2011, Dosovitskiy2015, Kulkarni2015}. The former constrains different timescale properties at different levels of the model network by assigning time constants, and the latter provides different spatial constraints at different levels by assigning specific local connectivity and local receptive field sizes. The current study examines how a spatio-temporal hierarchy adequate for the prediction of complex dynamic visual patterns can be developed by applying these two constrains to the activity in the P-MSTRNN model during the course of learning. The current study also examines how the same functional hierarchy developed through learning can be used in the recognition of current visual perception. 
The idea here is that an optimal intention state (one of the network’s latent states represented by activation patterns of context units in the RNN model) is inferred as that which ideally matches the perceived stream, as informed by the bottom-up prediction error signal. Finally, it is worth noting that spatial processing and temporal processing are performed simultaneously in the P-MSTRNN, whereas in other predictive coding models for dynamic vision processing they are not \citep{Srivastava2015}.  

 The P-MSTRNN performed a set of simulation experiments involving learning, predicting/generating, and recognizing visual human movement patterns. First, we video-recorded exemplars of human movement patterns. Human subjects were asked to generate cyclic body movement patterns by following particular movement syntax rules. Next, we scrutinized the developmental process of the neurodynamic structures in different layers in the model as it learned to predict movements within these exemplar patterns. Then, we examined the dynamics characteristic of active recognition (active inference) \citep{Friston2011}  during a predictive imitation (imitative synchronization) task \citep{Ahmadi2017} where test visual movement patterns were proactively imitated with synchrony by the model network. Informed by the prediction error signal, the model inferred intention states (latent states) corresponding to given test visual movement patterns. Finally, we looked at the generation and recognition of visual movement patterns during different stages of the learning process, in order to uncover possible relationships between the performance characteristics of the model network and its internal dynamic structure at each stage. Interestingly, transient dynamics - which develop before limit cycle attractors embedding target patterns develop - can themselves be used as memory for predicting and recognizing learned patterns.

\section{Model}
\subsection{Overview}
The predictive multiple spatio-temporal scales RNN (P-MSTRNN) develops a spatio-temporal hierarchy by extracting compositionality latent in exemplar dynamic visual streams. It is our conjecture that such hierarchy self-organizes when different spatial and temporal constraints are imposed simultaneously on neural activity in different layers of the network. 

The P-MSTRNN consists of a series of context layers, each composed of leaky integrator units \citep{Jaeger2007}.  The P-MSTRNN employs convolution as well as deconvolution operations as in convolutional (CNN) \citep{LeCun1998} and deconvolutional (deCNN) \citep{Zeiler2011, Dosovitskiy2015, Kulkarni2015} neural networks. This provides spatial constraints on neural information processing such that the lower levels preserve more local connectivity with smaller receptive fields while the higher levels preserve more global connectivity with larger receptive fields. However, unlike the CNN or deCNN, this model is capable of processing temporal information using recurrently connected leaky integrator neural units regulated by specific time constants. The leaky integrator integrates the history of its internal state and partially updates this state by receiving visual input from outside. In the leaky integrator, the time constant determines the portion of the internal state to be updated. In this way, the time constant constrains the timescale of neural activation dynamics. The larger the leaky integrator time constant value, the slower its internal state decays. Conversely, the smaller the leaky integrator time constant, the faster the internal state changes. 

Different timescales of different context layers preserve different dynamics. The lower layer is designated to preserve faster dynamics with a smaller time constant setting, and the higher layer preserves slower dynamics by way of a larger time constant. Following the operational principle adopted in \citet{Yamashita2008}, it is assumed that more abstract representation develops in higher layers, and more detailed in lower layers as the functional hierarchy forms.

In terms of the predictive coding scheme framing the model, current top-down intention propagates from higher context layers to lower context layers, and finally projects “intended” or predicted visual image patterns at the pixel level. Then, an error signal is generated between predicted and actually perceived pixel patterns. This signal is back-propagated from the lower layer to the higher ones, and activation values of context units at each layer are modified in the direction of minimizing this error. This bottom-up process of modifying all latent variables in all context layers corresponds to active recognition by way of error regression \citep{Tani1999} or active inference of intention \citep{Friston2005}. 

Various cognitive processes including recognition, generation and learning involve this dynamic process of top-down and bottom-up interaction \citep{Giese2013, Tani2016}. In the P-MSTRNN, learning involves optimizing connectivity weights in the whole network in the direction of minimizing the reconstruction error, using exemplar visual sequence patterns as the reconstruction targets. The learning process determines not only the optimal connectivity weights but also the corresponding latent states generative of exemplar patterns, represented by the initial states of all context units in all layers regenerating each learning target sequence. 
After the learning process converges, target visual sequence test patterns can be recognized by inferring latent states generative of target visual sequences with minimum error.

\begin{figure}
  \centering
  \includegraphics[width=1\textwidth]{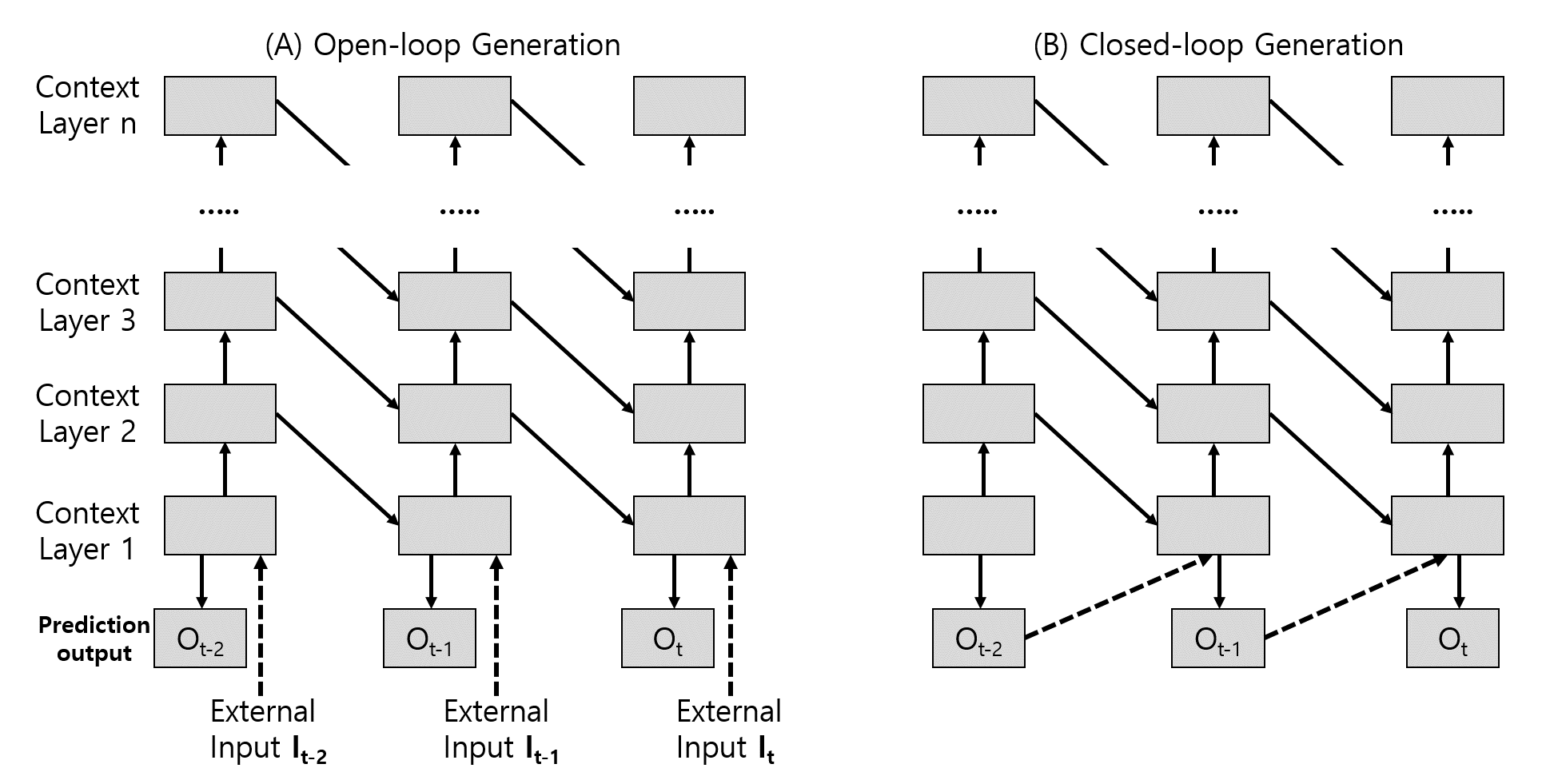}
  \caption{Conceptual schematics of two different output generation modes. (A) In the open-loop generation mode, the prediction outputs for the next steps are computed using current external visual input. (B) In the closed-loop generation mode, the prediction outputs for the next steps are computed by feeding prediction outputs from previous steps into current visual inputs without using external visual inputs. Schematics are presented in a succinct manner to address the essential differences between these two methods, omitting unimportant details.}
\end{figure}

Technically, the learned model network can be operated in three different modes, namely open-loop prediction, closed-loop prediction, and closed-loop prediction with error regression. During open-loop prediction, the model network predicts the visual input of the next step using the current visual input. During closed-loop prediction, the look-ahead prediction of multiple steps can be conducted, but not by using external visual inputs. Instead, prediction outputs from the previous step are fed into the visual inputs of the current step (see Fig.1 (A) (B)) \citep{Tani1996, Ziemke2005}. Generation of such internally generated image sequences can account for mental imagery \citep{Hesslow2002, Pezzulo2014}.

During closed-loop prediction using the error regression, multiple steps of a visual sequence are predicted using immediately past time steps. Inside this time window, the past perceptual sequence is reconstructed by closed-loop operation. Once the window-length closed-loop reconstruction is calculated, the error between the reconstructed sequence and the actual perception history is calculated and back-propagated through time inside the window. This back-propagated error then updates latent states so that the regenerated sequence inside the window best resembles actual (past) perception. 
The idea underlying this operation is that the future is predicted by quickly adapting latent states at multiple levels using the prediction error minimization scheme, in order to anticipate the ongoing perceptual flow as closely as possible \citep{Tani1999, Tani2003}. For example in \citet{Ito2004, Ahmadi2017}, model networks learned sets of visually perceived human movement patterns, and proactively imitated ongoing visual movement in effect performing predictive imitation using only the error regression. With predictive imitation by error regression, when the subject abruptly changes from one to another movement pattern, the model network is able to immediately read the change in a subject’s intention and to modify its generation of visual mental imagery patterns to match by applying the aforementioned error regression scheme to a window on the immediate past (see Fig.2). The next subsections detail this model.

\begin{figure}
  \centering
  \includegraphics[width=1\textwidth]{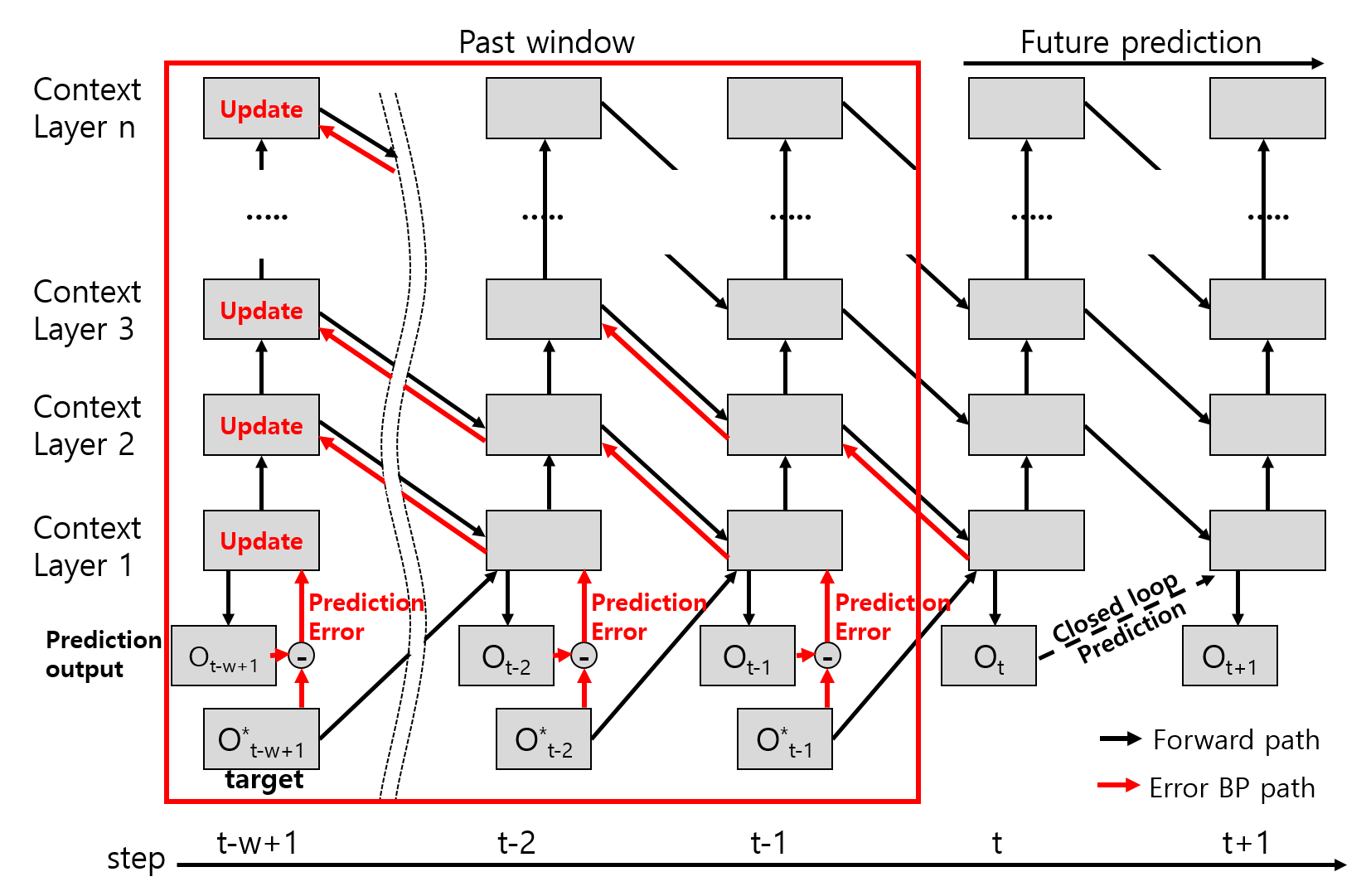}
  \caption{Schematic of predictive imitation by error regression. Black arrows indicate forward pathways and red arrows the error back-propagation through time pathway. Dotted arrows indicate closed-loop visual input. The red box is the window opening w steps into the immediate past. O*t is the target of time step t and Ot is the prediction output of step t. The prediction is produced closed-loop as indicated with dotted arrows. The error calculated between output and target sequences in the immediate past window is back-propagated through time steps. The back-propagated error is used to update all neural units in all context layers from the onset of the window in the direction of minimizing the error, and the connectivity weights and biases are fixed. By using the updated latent state at the onset of the window, the latent states in all steps forward in the window are re-computed by means of forward computation, along the solid black arrows shown in the figure. 
This latent state update is performed multiple times during every single video frame step forward. As the window advances to each next step, the process is performed again.}
\end{figure}

\subsection{Architecture}
The P-MSTRNN forward dynamical pathway is shown in Figure 3. Each context layer receives top-down and bottom-up signals from higher and lower levels of the network in order to calculate its own internal states. Specifically, the first layer receives bottom-up input from outside of the network as well as the top-down future prediction from inside of the network. The prediction error is calculated from the difference between these two, and is back-propagated. The backward process utilizes the same pathway as the forward process, but in the reverse direction. In both training and during recognition, the back-propagated error is used for updating network parameters such as connectivity weights and biases of all context layers, in effect adapting intention to anticipated input.

\begin{figure}
  \centering
  \includegraphics[width=0.7\textwidth]{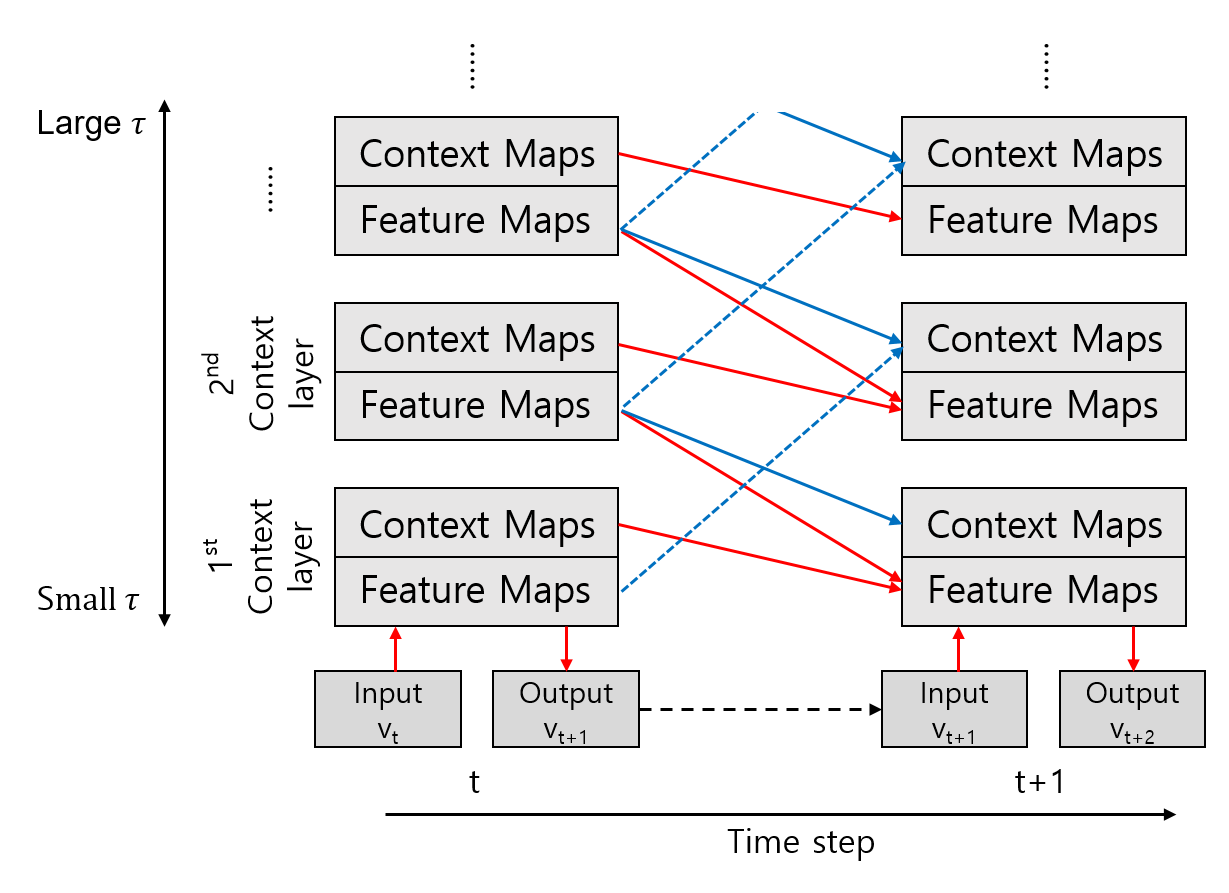}
  \caption{Structure of the P-MSTRNN model. The red arrow indicates the convolution operation, the blue arrow indicates element-wise multiplication and the blue dashed arrow is the bottom-up pathway. The black dashed arrow between the prediction output and visual input is the closed-loop path. Recurrent and leaky connections are not shown to keep the network structure clear. Once output is generated, error calculated between output and target is propagated backwards through the forward process.}
\end{figure}

The P-MSTRNN adds recurrency to key concepts from convolutional neural networks (CNN) \citep{LeCun1998} and deconvolutional neural networks \citep{Zeiler2011, Dosovitskiy2015, Kulkarni2015}. In the model, there are two basic units operating over the temporal dimension, feature units and context units. Both types are leaky integrator neural units. The internal state of each leaky integrator unit decays at each time step by a predefined amount, and it then receives inputs from outside. In addition to leaky integration, context units are recurrently connected, enabling strong temporal dynamics modulated by trainable recurrent parameters. 

In the model, a set of feature units builds a feature map (FM) and a set of context units forms a context map (CM). Both FMs and CMs are building blocks of the P-MSTRNN. FMs mostly deal with spatial processing and CMs contribute mostly to temporal dynamics. The timescales which determine each leaky integrator''s decay rate are set to small values in lower layers and to larger values in higher layers. The smallest time constant in the lowest layer is 2. Each upward layer has a time constant setting twice that of the prior lower layer. With these timescale settings, lower layer neural activity tends to deal with rapidly changing information and higher layer neural activity captures slowly changing information.

The forward dynamics of both FMs and CMs are given in Equations (1) to (4). 

\begin{equation}
    \bm{\hat{f}}^{lp}_t = \Big(1-\frac{1}{\tau^{l}}\Big)\bm{\hat{f}}^{lp}_{t-1} + \frac{1}{\tau^l}\Big(\sum_{q=1}^{Q_{l+1}}\bm{f}^{(l+1)q}_{t-1} * \bm{k}^{lpq}_{ff} +  \sum_{n=1}^{N_l}\bm{c}^{ln}_{t-1}*\bm{k}^{lpn}_{cf}
     + I*\bm{k}^{lp}_{if} + {b}^{lp}\Big)
\end{equation}

\begin{equation}
    \bm{f}^{lp}_t = 1.7159\tanh\Big(\frac{2}{3}\bm{\hat{f}}^{lp}_t\Big)
\end{equation}

where \(\hat{\bm{f}}\) and \(\hat{\bm{c}}\) indicate FM and CM internal states. \({\bm{f}}\) and \({\bm{c}}\) indicate FM and CM activation values. Subscripts for internal state and activation value indicate time steps, and superscripts indicate the location of a particular FM or CM in the model. For example, in Equation (1), \(\hat{\bm{f}}^{lp}_t\) indicates the internal state of the \(p^{th}\) FM in the \(l^{th}\) layer at time step t. Likewise, \(\bm{c}^{ln}_{t-1}\) indicates the activation value of the \(n^{th}\) CM in the \(l^{th}\) layer at time step t-1. The network parameter \(\bm{k}\) indicates kernel weight and \(\bm{b}\) indicates bias. In greater detail, for example the subscript in \(\bm{k}_{ff}\) indicates that the kernel connects an FM to an FM. The superscript \(ff\) in \(\bm{k}^{l}_{ff}\) indicates that the kernel always connects the previous time step’s FM in layer (l+1) to the current step’s FM in \(l^{th}\) layer. The superscript in kernel \(\bm{k}^{lpq}_{ff}\) indicates that the \(q^{th}\) FM in layer l+1 at the previous time step t-1 is linked to FM p in layer l at the current time step t through the kernel. \(\bm{k}^{lpn}_{cf}\) indicates that the kernel connects the \(n^{th}\) CM in the \(l^{th}\) layer at the previous time step to the \(p^{th}\) FM in the \(l^{th}\) layer at the current time step t. \(b^{lp}\) indicates the bias of the \(p^{th}\) FM or CM in the \(l^{th}\) layer. 

Equation (1) expresses how the internal states of a FM is formed. The first term indicates the internal state of the FM as it has decayed from the previous time step. The time constant \(\tau^l\) of the \(l^{th}\) layer determines the decay rate for that layer according to the ratio \((1-\frac{1}{\tau^l})\). The second term is an input signal to the current FM from the activation values of previous time step FMs in the \(l+1^{th}\) layer. \(Q_{l+1}\) is the number of FMs in the \(l+1^{th}\) layer and the operator * is a convolution operation. \(N_{l}\) is the number of CMs in the lth layer. The fourth term indicates input from outside of the network. This term only appears when the current layer is the first layer. Other levels take input from inside of the network, and this term disappears. We used the scaled hyperbolic tangent activation function suggested by \citet{LeCun2012} for efficient back-propagation in the model  (Equation (2)). 

Equation (3) and (4) detail the forward dynamics of CMs. 

\begin{multline}
    \bm{\hat{c}}^{lm}_t = \Big(1-\frac{1}{\tau^{l}}\Big)\bm{\hat{c}}^{lm}_{t-1} + \frac{1}{\tau^{l}}\Big(\sum_{n=1}^{N_l}\bm{c}^{ln}_{t-1}\odot\bm{W}^{lmn}_{cc} + \sum_{q=1}^{Q_{l+1}}\bm{f}^{(l+1)q}_{t-1}\odot\bm{W}^{lmq}_{fc}\\ + 
    \sum_{r=1}^{R_{l-1}}\bm{f}^{(l-1)r}_{t-1}*\bm{k}^{lmr}_{fc}+{b}^{lm}\Big)
\end{multline}

\begin{equation}
    \bm{c}^{lm}_t = 1.7159 \tanh\Big(\frac{2}{3}\bm{\hat{c}}^{lm}_t\Big)
\end{equation}    

In Equation (3), CM internal states also utilize leaky integrator dynamics with a decay rate of \((1-\frac{1}{\tau^l})\). The second term indicates input from the previous step'’s CM in the same layer. \(\bm{W}^{lmn}_{cc}\) is the weight of the recurrent connection between the previous step’s \(n^{th}\) CM to the current step’'s \(m^{th}\) CM in the same layer. The operation \(\odot\) is the element-wise multiplication, with both sides of this operator always the same size. For example, in Equation (3), \(\bm{c}^{ln}_{t-1}\) and \(\bm{W}^{lmn}_{cc}\) have the same size and the \(\odot\) operation’s result is also the same size. The third term is the input from the previous step’s FM in the \(l+1^{th}\) layer. The fourth term indicates FM input from the \(l-1^{th}\) layer. Kernel \(\bm{k}^{lmr}_{fc}\) connects the \(r^{th}\) FM in the \(l-1^{th}\) layer to the current \(m^{th}\) CM. When calculating the convolution operation, if the size of the input map is smaller than that of the output, zero-padding of the input is used because it is the cheapest way in terms of computation without meaningful degradation of performance. 

	Lastly, prediction output is obtained from Equations (5) and (6). 

\begin{equation}
    \bm{\hat{O}}_t = \sum_{q=1}^{Q_1}(\bm{f}^{1q}_t*\bm{k}^q_{fo} + {b}_o)
\end{equation}
    
\begin{equation}
    \bm{O}_t = 1.7159\tanh\Big(\frac{2}{3}\bm{\hat{O}}_{t}\Big)
\end{equation}

where \(\hat{\bm{O}}_t\) is obtained from the values of the first FM at the same time step. The convolution operation and activation function are the same as those described above. 

\subsection{Learning and generation methods}
During training, after prediction output is obtained, the difference between the output frame and the prediction target frame at time step t is calculated, resulting in error for all time steps. This process is shown in the Equation below. 

\begin{equation}
    E = \frac{1}{T}\sum_{t=1}^{T}E_t
\end{equation}

\begin{equation}
    E_t = \frac{1}{XY}\sum_{i}^{X}\sum_{j}^{Y}(O^{*}_{ij}-O_{ij})^2
\end{equation}

where \(E_t\) is the average error per pixel at time step t and \(E\) s the average error per pixel for total time steps T. \(\bm{O}^*_{ij}\) is the target pixel value in the (i,j) position of the frame and \(\bm{O}_{ij}\) s the output pixel in the (i,j) position of the frame. The target  \(\bm{O}^*_{ij}\) can be multiple time-steps ahead. Output produced by the current step t can predict one-step-ahead, two-steps-ahead and further. In this study, we used two-steps-ahead prediction, meaning that output at time step t is the prediction of step t+2. We empirically found that two-steps-ahead prediction facilitates faster training than one-step-ahead prediction, conjecturing that this is due to there being bigger differences between current input and a target two time steps ahead than between current input and one step ahead. This bigger difference yields larger gradients meaning faster training. The error calculated through this process is then used to optimize network parameters. A conventional back-propagation through time (BPTT) method whereby parameters such as weights, kernels, biases and initial states of the layers (latent states) are optimized using gradient descent.

The network can be trained using open-loop generation and closed-loop generation. In this paper, both open-loop generation and closed-loop generation train the network at the same time. For each training epoch, a closed-loop generation is performed first, generating prediction output without adjusting parameters. Closed-loop prediction output is then mixed with data frames corresponding to each time step, which are used as inputs to the network being trained. The mixing ratio of data frames to closed-loop outputs is \(\alpha : 1-\alpha \). With the closed-loop output at time step t-1 equal to \(O^{t-1}_{close}\) and the data frame at t equal to \(I^{t}_{data}\), the mixed input frame at time step t is expressed as \((1-\alpha)O^{t-1}_{close} + \alpha I^{t}_{data}\). \(\alpha\) is 0.9 in this paper. If \(\alpha\) is 1, only open-loop generation is used for training, and if 0 then only closed-loop generation is used instead. This mixed version of open-loop and closed-loop generation allows for faster training.

During training, error from the output layer is used to optimize network parameters. As training proceeds, both open-loop error and closed-loop error decrease. In most cases, closed-loop error is higher than open-loop error because closed-loop prediction generates accumulated error over time steps without correction from factors outside of the network. On the other hand,  open-loop generation is driven by external data frames which inform internal states and correct prediction outputs. In the current research, network training is terminated when closed-loop error reaches a predefined lower bound threshold at which point the network is guaranteed to successfully perform both open-loop generation and closed-loop generation. 

All recurrent type neural networks form internal states of the current time step partly by receiving signals from the previous step. At the first step however, there is no previous step. Therefore, initial state values must be formed. In order for the network to learn to generate multiple data sequences, it must infer optimal initial states of all context units in all layers for each sequence. As with the optimization of connectivity weights, inference of optimal initial states is performed via error BPTT. All training sequences have their own optimal initial states, and those inferred initial states for each training sequence represent intention within the context of that specific sequence. Therefore in the generation process, the initial state corresponding to a target sequence that we want to generate, as an ideal toward which the network then aims, must be fed to the network at the beginning.

\subsection{Inferring intention by error regression during predictive imitation}
The trained network both generates and recognizes learned sequence patterns. A sequence pattern can be generated by setting the latent state consisting of the initial states of all context units to their corresponding values.
On the other hand, in predictive imitation by error regression given a target visual input sequence, an optimal latent state minimizing reconstruction error must be generated.
For this purpose, the idea of the error regression window \citep{Tani2003, Murata2015, Ahmadi2017} is adopted to the current model. 
In the regression window, forward and backward computations are iterated within the immediate temporal window of the past w steps, updating the latent state in the direction of minimizing error.
The latent state consists of the values of all context units in all layers at the onset of the error regression window (See Fig.2).
The detailed procedure is presented in Table 1. \textit{AllTimeSteps} specifies that the process will be iterated for this number of steps. For each step during the process, the model receives visual input \(V_t\) and saves it into the input history buffer \textit{R}. Then, it initializes the prediction error value inside the temporal window as 0. Then, the network produces closed-loop prediction \textit{P} inside the temporal window. For \textit{TemporalWindowSize} \textit{W}, starting from \(t-W+1^{th}\) to \(t^{th}\) step (from \(P_{t-W+1}\) to  \(P_t\)), prediction error within the temporal window is calculated by comparing predictions (\textit{P}) to the target (\textit{V}) from input history buffer \textit{R}. If the error value is bigger than the threshold, this prediction error is back-propagated through time and used to modify the initial states within all steps of the temporal window in the direction of producing correct prediction thus minimizing error. On the contrary, if the error value is smaller than the threshold, it means the network is producing acceptable synchronized prediction for the current step \(t\). Finally, the process of error regression proceeds to the next step \(t+1\) and the cycle repeats itself. 

\begin{table}[]
\centering
\caption{Algorithm for Error regression}
\label{my-label}
\begin{tabular}{|l|}
\hline
Error regression algorithm                                                                                                                            \\ \hline
\begin{tabular}[c]{@{}l@{}}
\textbf{for} t in \textit{AllTimeStep}: \\
\quad \(Receive\_visual\_input\_V_{t}\) \\
\quad \(Add\_visual\_input\_into\_input\_history\_buffer\_R\) \\
\quad \textbf{while(true)}:\\
\quad\quad \(E_{window} = 0\) \\
\quad\quad \textbf{for} w in \textit{TemporalWindowSize} (=W): \\
\quad\quad\quad \(Generate\_prediction\_P_{t-W+w}\) \\
\quad\quad\quad \(Calculate\_error\_E_{t-W+w}\)\\
\quad\quad\quad \(E_{window} = E_{window} + E_{t-W+w}\)\\
\quad\quad \textbf{if} \(E_{window} < threshold:\)\\
\quad\quad\quad \textit{break} \\
\quad\quad \textbf{else}:\\
\quad\quad\quad \(Backpropagate\_error\_within\_Temporal\_Window\)\\
\quad\quad\quad \(Modify\_initial\_states\_in\_Temporal\_Window\)\\
\quad \(Generate\_Prediction\_for\_current\_step\_P_{tf}\)

\end{tabular} \\ \hline
\end{tabular}
\end{table}

\section{Experiments}
We conducted a set of simulation experiments investigating the dynamics of the P-MSTRNN using a video image dataset composed of whole-body movement patterns performed by multiple human subjects. This set of movement patterns was organized according to a pre-defined syntax for generating body movement patterns, described in detail in section 3.1. 

Experiment 1 examined how memory dynamics develop by self-organizing hierarchical structures. For this purpose, we observed temporal neuro-dynamic structures at different stages of learning and compared them to each other. Additional study investigated more complex sequence patterns consisting of concatenations of primitive segments to reveal how the trained network utilizes its functional hierarchy to deal with complex sequences. 

Experiment 2 investigated possible relationships between dynamical structure development and task performance. We assessed  active recognition capability in terms of predictive imitation by error regression at different stages of training.  We also compared predictive imitation performance using two distinct mechanisms, namely predictive coding by error regression and ordinary entrainment by input, for example as in \citet{Kelso1997, Taga1991}.

\subsection{Dataset}
The dataset used for the experiments consists of multiple sequences, each of which containing only one movement primitive pattern. 

\begin{figure}
  \centering
  \includegraphics[width=0.7\textwidth]{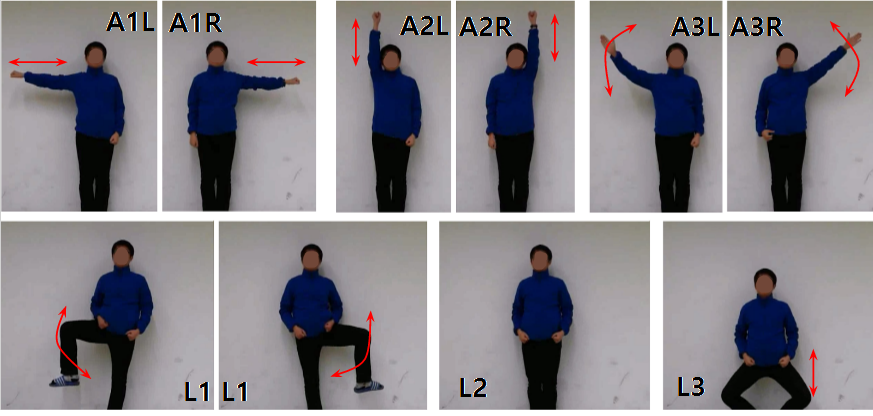}
  \caption{The sub-primitives of legs and arms. There are three arm movement and three leg movement sub-primitives. These sub-primitives are combined to constitute the set of whole-body movement primitives. }
\end{figure}

The set of primitive movement patterns consists of six types of whole-body movement patterns as demonstrated by five different subjects and collected by video shooting performances. We first designed sub-primitives of legs and arms, then composed the whole-body movement primitives using sub-primitives. Figure 4 shows leg and arm sub-primitives. For the arm, there are three sub-primitives. Arm sub-primitive 1 (A1) is extending arms horizontally. (A2) is vertically extending arms. (A3) is drawing big circles with laterally extended arms. We labeled sub-primitives for each side of the arms. Sub-primitive 1 of the left arm is labeled A1L, and of the right arm A1R. Similarly, sub-primitive 2 of the left arm is labeled A2L, the sub-primitive 2 of the right arm A2R, and sub-primitive 3 of left arm A3L and the sub-primitive 3 of right arm A3R. For the leg sub-primitives shown in the second row of Figure 4, there are also three types. Sub-primitive 1 of legs is raising the legs (L1) and sub-primitive 2 of legs is standing still (L2). Sub-primitive 3 of legs is bending legs (L3). These sub-primitives are labeled according to side (as were the arm sub-primitives, above). The syntax of composing whole-body primitives is presented in Table 2.

\begin{table}[h]
\centering
\caption{Hierarchical syntax of action primitives}
\label{my-label}
\begin{tabular}{@{}ccccccccccccc@{}}
\toprule
                     & \multicolumn{2}{c}{P1}       & \multicolumn{2}{c}{P2}         & \multicolumn{2}{c}{P3}       & \multicolumn{2}{c}{P4}         & \multicolumn{2}{c}{P5}       & \multicolumn{2}{c}{P6}         \\ \midrule
\multirow{3}{*}{Arm} & Left         & Right         & Left          & Right          & Left         & Right         & Left          & Right          & Left         & Right         & Left          & Right          \\ \cmidrule(l){2-13} 
                     & A2L          & A1R           & A1L           & A2R            & A3L          & A3R           & A3L           & A3R            & A1L          & A1R           & A2L           & A2R            \\
                     & \multicolumn{2}{c}{Co-phase} & \multicolumn{2}{c}{Anti-phase} & \multicolumn{2}{c}{Co-phase} & \multicolumn{2}{c}{Anti-phase} & \multicolumn{2}{c}{Co-phase} & \multicolumn{2}{c}{Anti-phase} \\ \midrule
Leg                  & \multicolumn{2}{c}{L1}       & \multicolumn{2}{c}{L2}         & \multicolumn{2}{c}{L1}       & \multicolumn{2}{c}{L2}         & \multicolumn{2}{c}{L3}       & \multicolumn{2}{c}{L3}         \\ \bottomrule
\end{tabular}
\end{table}

In Table 2, the term "co-phase" indicates that two arm actions should be performed at the same time, while "anti-phase" means that arm actions should be performed alternatively. As shown in the syntax table, each sub-primitive is utilized twice. Each sequence of whole-body action primitives used in both the training and recognition datasets consists of six cycles of each whole-body primitive, around 150 steps. 
Figure 5 shows the six movement primitives used for training as exemplified by a single human actor. 

\begin{figure}
  \centering
  \includegraphics[width=0.7\textwidth]{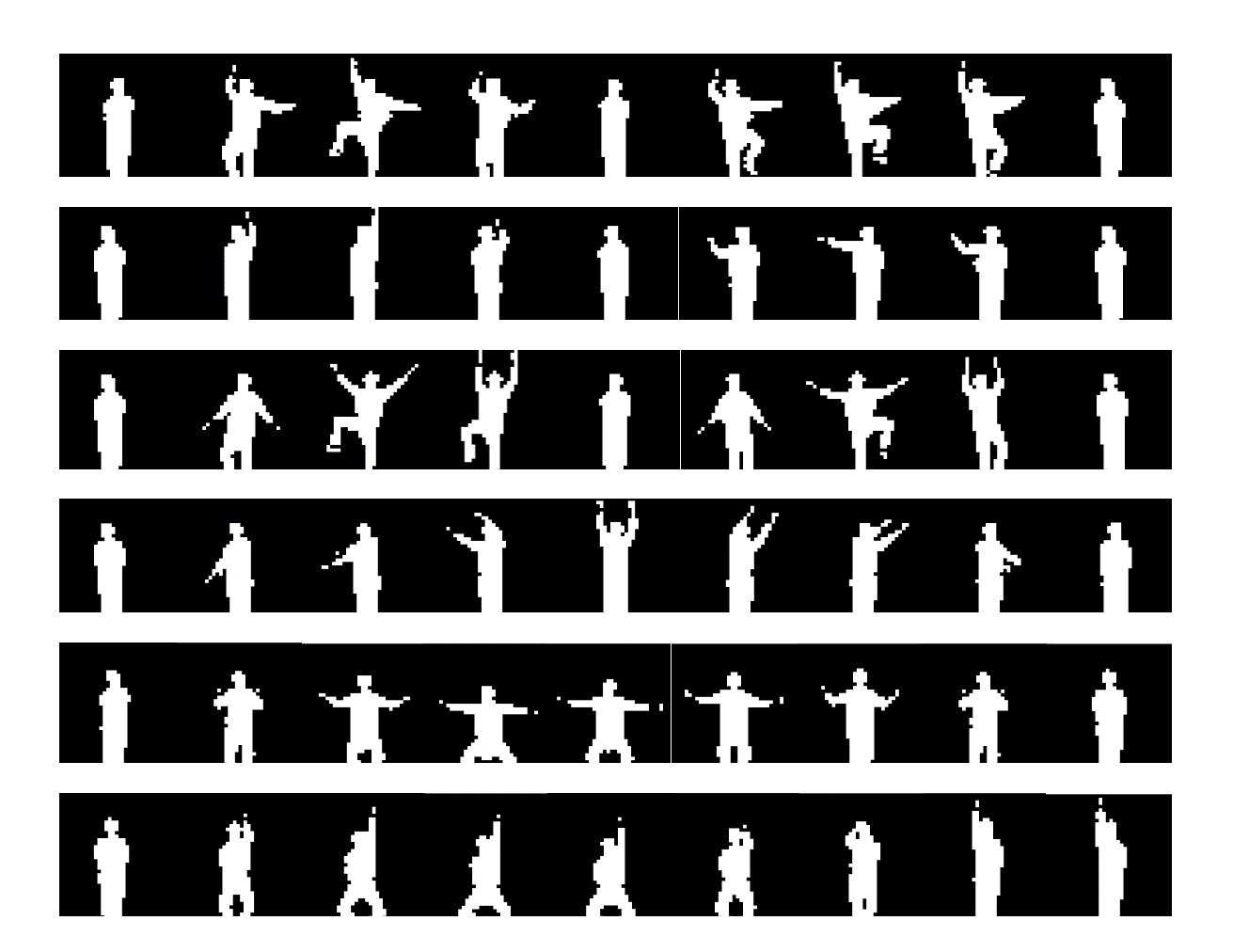}
  \caption{
  Examples of training data. Six types of whole-body movement primitives (P1, P2, P3, P4, P5 and P6) as demonstrated by the same person are plotted from the top row, down -  P1, P2, P3, P4, P5 and P6.}
\end{figure}

\subsection{Network}
The same P-MSTRNN model structure is used throughout this paper. Table 3 shows the network structure used for all experiments. The 0th layer is the input/output layer and the rest of the layers are context layers. Local connectivity and kernel sizes are defined by local maps. The learning rate was set to 0.002 for experiment 1 and 0.1 for experiment 2.

\begin{table}[h]
\centering
\caption{Parameter size of the network used for experiments }
\label{my-label}
\begin{tabular}{ccccccccc}
\hline
      & \multicolumn{4}{c}{feature map}                                                                                                   & \multicolumn{4}{c}{context map}                                                                                                   \\ \hline
layer & \begin{tabular}[c]{@{}c@{}}time \\ constant\end{tabular} & size  & number & \begin{tabular}[c]{@{}c@{}}kernel\\ size\end{tabular} & \begin{tabular}[c]{@{}c@{}}time \\ constant\end{tabular} & size  & number & \begin{tabular}[c]{@{}c@{}}weight\\ size\end{tabular} \\ \hline
0     & 1                                                        & 36,36 & 1      & 5,5                                                   & -                                                        & -     & 0      & -                                                     \\
1     & 2                                                        & 32,32 & 10     & 7,7                                                   & 2                                                        & 26,26 & 10     & 26,26                                                 \\
2     & 4                                                        & 26,26 & 10     & 7,7                                                   & 4                                                        & 20,20 & 10     & 20,20                                                 \\
3     & 8                                                        & 20,20 & 20     & 9,9                                                   & 8                                                        & 12,12 & 10     & 12,12                                                 \\
4     & 16                                                       & 12,12 & 40     & 11,11                                                 & 16                                                       & 2,2   & 25     & 2,2                                                   \\
5     & 32                                                       & 2,2   & 10     & 2,2                                                   & 32                                                       & 1,1   & 10     & 1,1                                                   \\
6     & 64                                                       & 1,1   & 10     & 1,1                                                   & 64                                                       & 1,1   & 5      & -                                                     \\ \hline
\end{tabular}
\end{table}

\subsection{Experiment 1 - Settings}
Experiment 1 examined spatial-temporal hierarchy development in the P-MSTRNN  as it learned to predict human movements as demonstrated in the exemplar video images of human movement patterns described previously (Section 3.1: Dataset). In experiment 1a, the network first learned (for 8000 epochs) six visually perceived primitive movement patterns, each performed by five subjects (for a total of 30 sequences) . All training sequences and all corresponding network parameters including connectivity weights, biases, and initial states were saved at 6 different stages of training for comparison with structures developed during more complicated learning tasks, in experiment 1b. 

\subsubsection{Experiment 1a - Developmental processes of dynamic memory structures}
After 8000 epochs, the model network was tested for closed-loop visual sequence generation. All 6 different primitive patterns were regenerated\footnote{Supplementary Video1 at \\ \url{sites.google.com/site/academicpapersubmissions/p-mstrnn }}. Examples of closed-loop generation after 8000 epochs of training are shown in Figure 6. The regeneration of three of the movement patterns, P1, P2 and P5 appear in the first, second and the third rows, respectively. It was furthermore found that subject-wise differences in demonstrating the same primitives were also reconstructed to some degree. Detailed analysis on this point is given after analysis of the development process.

\begin{figure} 
  \centering
  \includegraphics[width=0.7\textwidth]{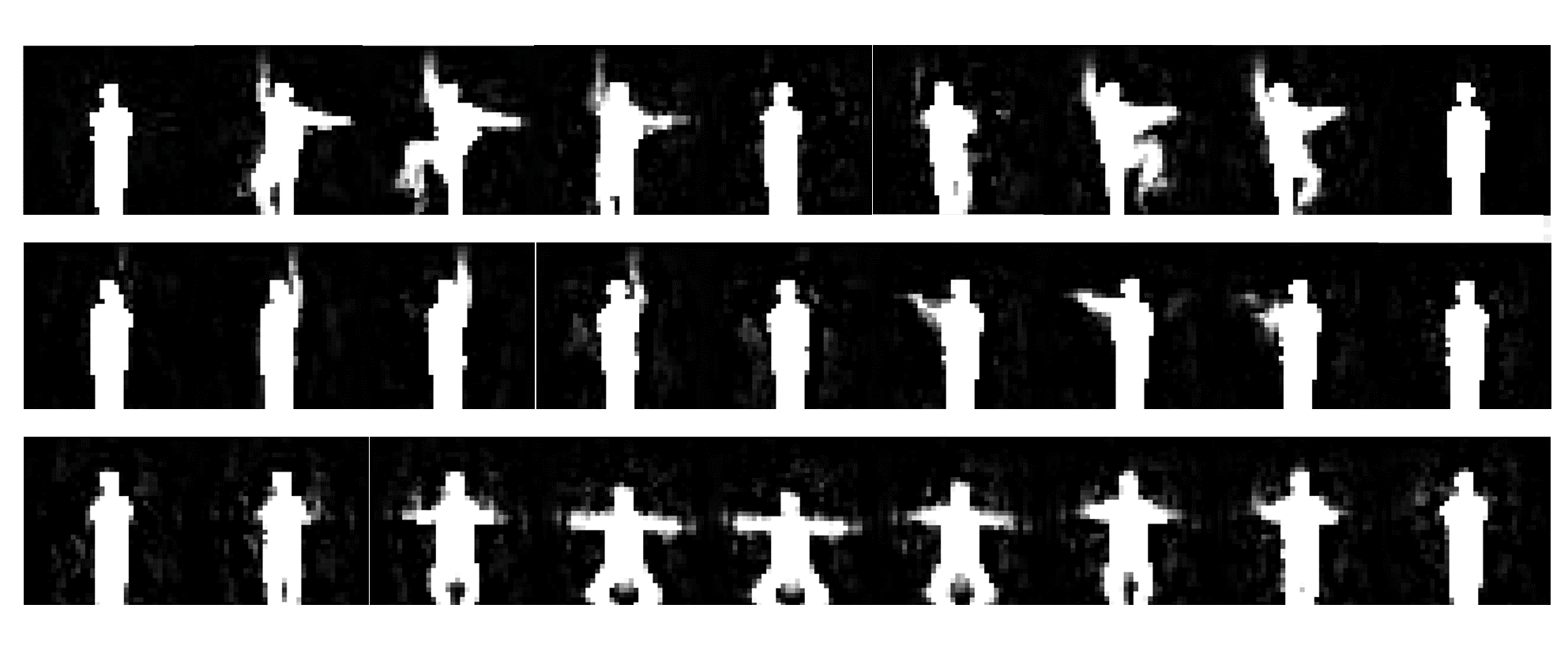}
  \caption{Examples of closed-loop generation of learned primitive movement patterns. P1, P2 and P5 from the top to the bottom. Closed-loop generation was performed by the network after 8000 training epochs.}
\end{figure}
We then analyzed neural trajectories at six different learning stages: at the \(100^{th}\) epoch, \(500^{th}\) epoch, \(1000^{th}\) epoch, \(2000^{th}\) epoch, \(4000^{th}\) epoch and \(8000^{th}\) epoch. We recorded the activities of all neural units during closed-loop regeneration at these stages. 
The neural trajectories were obtained at each stage of training by means of performing closed-loop generation for 1000 steps. The closed-loops started corresponding to the initial states adapted for target patterns. Then, we examined the neural trajectories for whether attractors of limit cycles or fixed points were formed. We used Principal Component Analysis (PCA) to reduce the number of dimensions describing the neural activity. The neural trajectories resulting from PCA applied to neural activities at different stages are shown in Figure 7. If the trajectory of neural activity presented in lower dimensions forms a limit-cycle attractor in later steps (\(700^{th}\) \textasciitilde \(1000^{th}\) steps) and if the pixel level generation maintains its corresponding visual pattern, this neural activity is counted as a limit-cycle attractor. However, if the visual pattern is broken or the trajectory does not form attractor, it is not.

\begin{figure}
  \centering
  \includegraphics[width=0.85\textwidth]{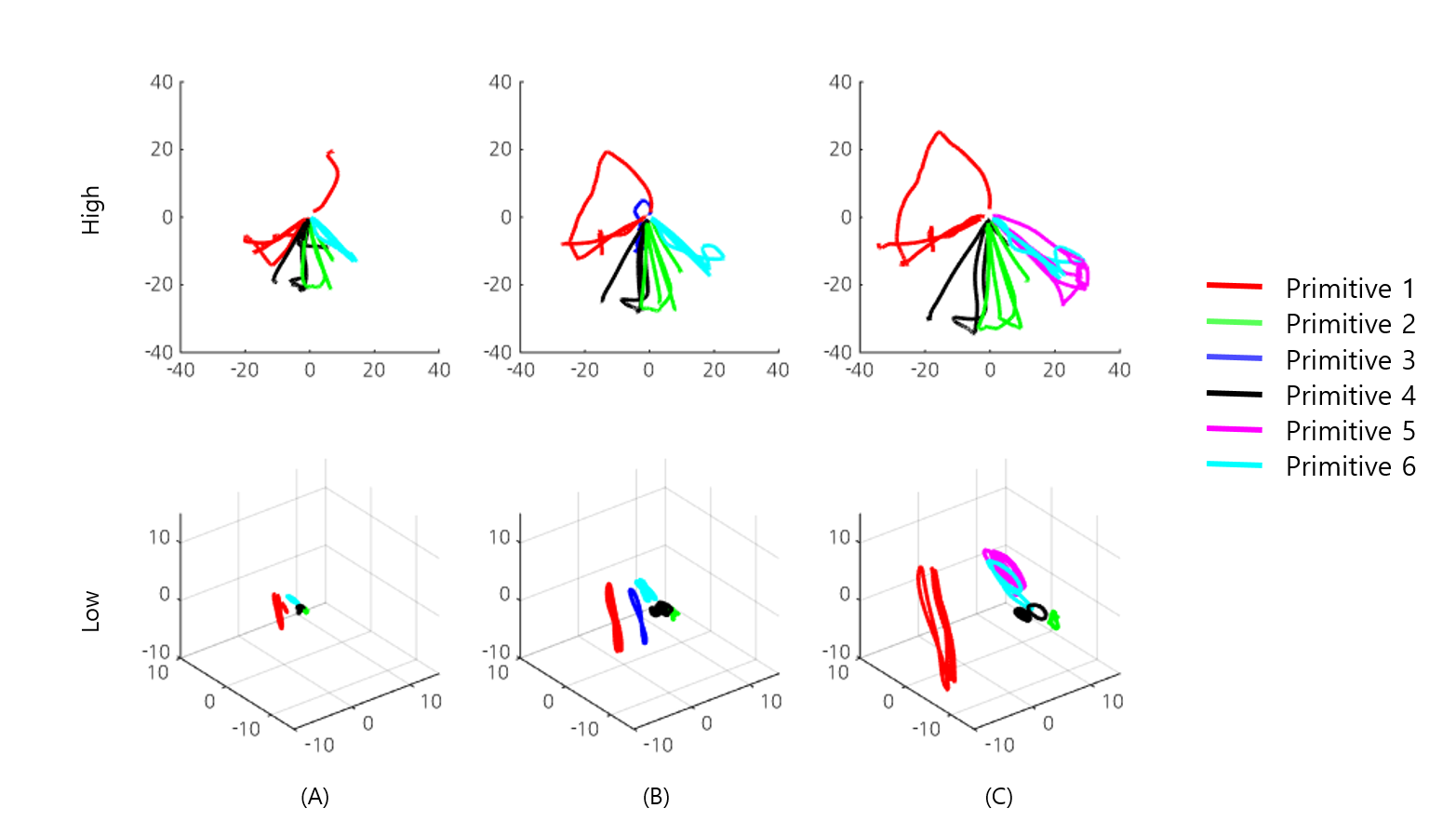}
  \caption{Neural trajectories generated in the higher (first row) and the lower (second row) layers. (A) The neural trajectories generated at 2000 epochs of training, (B) 4000 epochs, and (C) 8000 epochs. The neural trajectories shown in the same color belong to the same categories of primitive movement patterns as demonstrated by 5 different subjects each with his/her own characteristic differences.}
\end{figure}

In figure 7, neural trajectories from a higher layer (\(4^{th}\) layer FM) appear in the first row and neural trajectories from a lower layer (\(1^{st}\) layer CM) appear in the second row. Trajectories plotted in the same color belong to the same primitive movement pattern (P1-Red, P2-Green, P3-Blue, P4-Black, P5-Magenta and P6-Cyan) as demonstrated by 5 different subjects. Data from the \(2000^{th}\) training epoch, \(4000^{th}\) epoch, and \(8000^{th}\) epoch are shown in (A), (B) and (C) respectively.

The neural trajectories in the higher level were plotted from the first step to the \(1000^{th}\) step. All neural trajectories reached fixed points by the end of closed-loop generation. Neural trajectories began from the center region and ended in peripheral regions where fixed point attractors were located. Interestingly, higher layer neural trajectories for the same actions with the same colors demonstrated by different subjects reached neighboring fixed points, thus forming clusters. One interpretation of this higher layer clustering is that they represent intentions to activate primitive movement patterns. Neural trajectories in the lower layer were plotted from the \(900^{th}\) step to the \(1000^{th}\) step in order to examine limit cycle attractors in the absence of transient neural trajectories, and the different shapes of cyclic neural trajectory shown in different colors were observed after 1000 epochs of training.

Our interpretation of these results is that more abstract representations develop in higher layer due to the slower timescale constraints. On the other hand, more detailed information about the visual input sequences was captured in the lower layers due to the faster timescale constraints. By this thesis, the network developed its temporal hierarchy through learning due to the interaction between the multiple scales of temporal constraint imposed on neural activation dynamics in different layers of the model network. As the result, the different fixed points in the higher layer provide different bifurcation parameter values to the lower layer, allowing for the proactive selection among different limit cycle attractors corresponding to different learned movement patterns.

Table 4 shows the number of limit-cycle attractors at certain training epochs and that they appear even in the early stage of learning (1000th training epoch). Also, the trajectories of limit-cycle attractors formed during late steps are different from those of early steps. During early steps of generation, trajectories are not stable and there is a good amount of fluctuation, because the training patterns themselves are of human movement with some variations. However, as generation goes on, the trajectories become more stable in terms of magnitude and length of cycle. 

The fact that a network trained with fluctuating patterns eventually generates neural activity forming a regular limit-cycle attractor implies that the network has the ability to generalize its training pattern. If the network were only memorizing a pattern, fluctuations in cycle length and pattern magnitude during early steps would be replicated also in later steps. Moreover, if the model were simply memorizing the training sequences, it would not generate closed loop prediction for steps longer than the training sequences.

The fact that stable limit-cycle attractors form during relatively early learning, at around the 1000th epoch, implies that the network is not merely memorizing the pattern by over-iteration of training. Also, limit-cycle attractors (which are a generalized form of training patterns) would not appear.
\begin{table}[h]
\centering
\caption{The average number of limit-cycle attractors embedding target movement patterns at each stage of training}
\label{my-label}
\begin{tabular}{ccccccc}
\hline
\begin{tabular}[c]{@{}c@{}}Training  Epoch\end{tabular}                                             & 100     & 500     & 1000    & 2000    & 4000    & 8000    \\ \hline
\begin{tabular}[c]{@{}c@{}}Learning MSE\end{tabular}                                               & 0.03779 & 0.03117 & 0.02254 & 0.01777 & 0.01397 & 0.01164 \\
\begin{tabular}[c]{@{}c@{}}Limit-cycle Attractors \\embedding target\end{tabular} & 0       & 0       & 4.5     & 13.75   & 13.5    & 15      \\ \hline
\end{tabular}
\end{table}

Considering the probabilistic nature of the development processes, we counted the number of limit-cycle attractors embedding targets three times, each time using the same structure of networks seeded with different random initial parameters from which learning then proceeded. Averages were calculated for the number of the attractors at each stage. 
The formation of limit-cycle attractors during later steps of generation (from 700 to 1000 steps) are plotted in the second row of figure 6. The neural activity of earlier steps (from 1 to 700 steps) is excluded as there are some fluctuations in terms of length of cycle and magnitude of pattern because the transient region reflects irregularities in training patterns.


Table 4 and Figure 7 clarify the developmental process. According to the table, during early learning (such as at the \(100^{th}\) and \(500^{th}\) epochs), the network has yet to form limit cycle attractors. After these early stages, the number of limit cycle attractors increases. Also, as training proceeds, the sizes of the lower level''s limit-cycle attractors increase and the distances between different attractors belonging to different primitive movement patterns increase as well (see the second row in Figure 7). A similar observation can be made for fixed point attractor positions in the higher layer. As learning proceeds, the positions of fixed point attractors expand outwards. This tendency for the number of the attractors to increase and to expand over the course of development implies that more detailed information about target patterns is preserved in the dynamic structures of the whole network as training proceeds. 

As noted above, the regeneration of the subject-wise idiosyncrasies in demonstrated movement patterns within the same categories can now be explained. For example, as the plots of the lower level at the \(8000^{th}\) epoch in Figure 7 (C) show, two distinct attractors belonging to movement primitive pattern 1 (red) develop. Let us name the right-hand side attractor ‘attractor 1’ and the left-hand side attractor ‘attractor 2’.

Figure 8 shows the reconstructed images corresponding to two distinct attractors from Figure 7 (C). 
\begin{figure}
  \centering
  \includegraphics[width=0.99\textwidth]{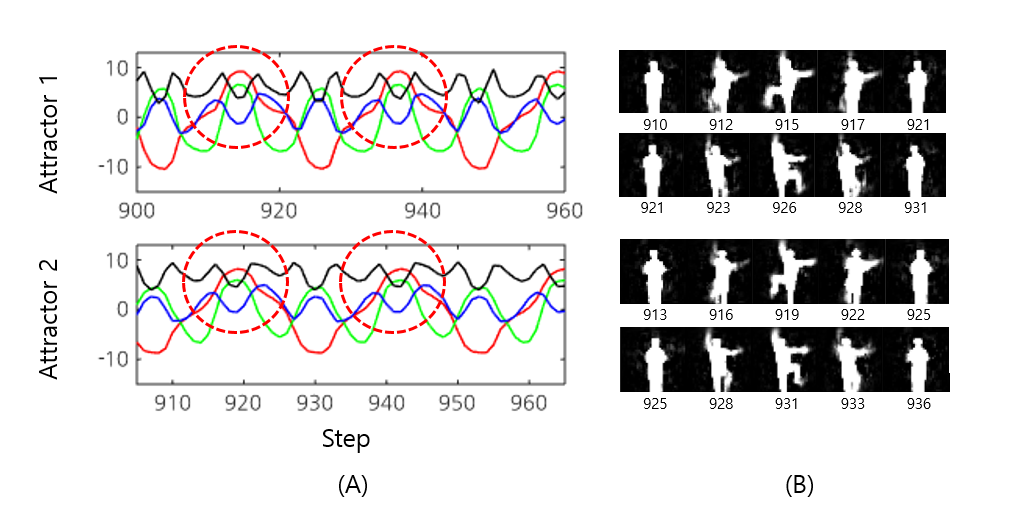}
  \caption{Closed-loop generation of two distinct patterns belonging to the same category as demonstrated by two different subjects. (A) PCA plots for attractor 1 and attractor 2. (B) Pixel level closed-loop generation of both attractors. Numbers indicate frame steps for the same action performed by different subjects. Marked positions show slight differences. }
\end{figure}
Figure 8 (A) shows the PCA plots of attractor 1 and attractor 2 in the upper and lower rows, respectively. These two plots look similar since they are from the same primitive movement pattern. After close inspection however, some differences can be seen. Marked regions indicate distinguishable features between attractors 1 and 2. Figure 8 (B) presents regenerated images at the pixel level for both attractors. The first and the second rows show snapshot images of attractor 1 and attractor 2, respectively. The pixel level images are different in their shape and size. The body shape in the image generated from attractor 1 seems slimmer than the one from attractor 2. Also, poses were different especially in regions of legs and arms. These observations imply that detailed information concerning subject-wise idiosyncrasies in the same category of primitive movement pattern is well preserved in adjacent attractors with similar shapes especially in later training stages. Indeed, these two distinct attractors are not identifiable in the plot of the previous training stage, for example as shown in Figure 7 (B) where we can see only one red-colored attractor.

As described previously, there are no limit-cycle attractors embedding target patterns after 500 epochs of training. However, although the target patterns could not be generated as stable limit-cycle attractors, each target pattern could be generated adequately from the corresponding initial state for at the least a few cycles. In order to examine this phenomenon more closely, 1000 steps of closed-loop generation were performed with the network at this stage. Figure 9 compares the closed-loop generation of the model network during early and late stages of training.

\begin{figure}
  \centering
  \includegraphics[width=0.99\textwidth]{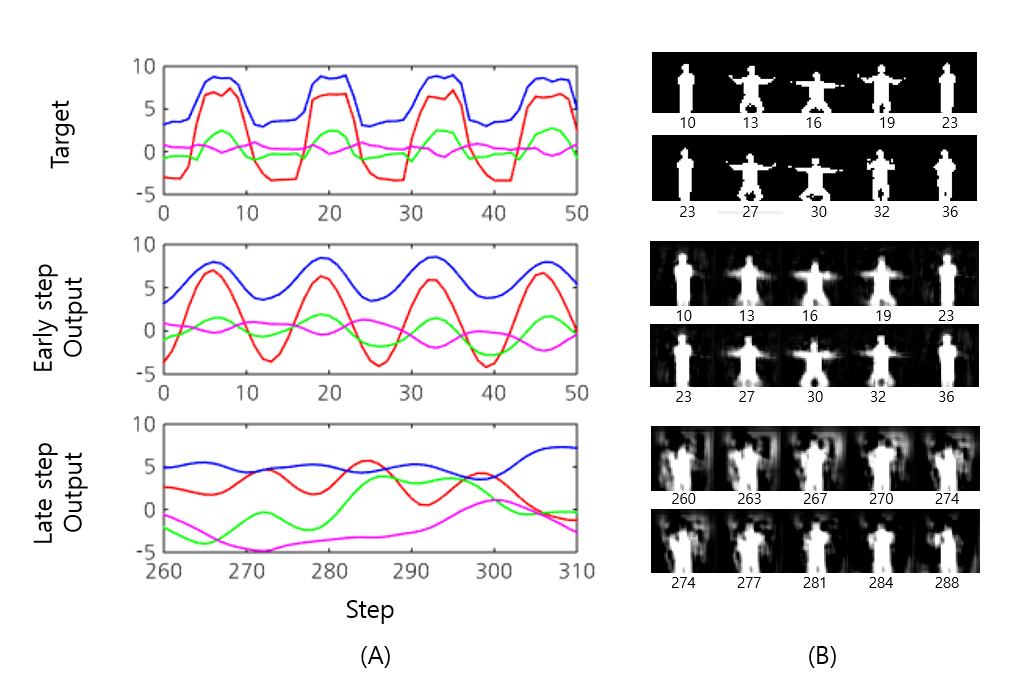}
  \caption{Closed-loop generation during the \(500^{th}\) epoch. The data for this figure are collected from the same network parameter trained for \(500^{th}\) epoch. The first row shows the target pattern in the early steps. The second row shows closed-loop output generation during the same period. The third row shows closed-loop output during later steps. 
  For each row, the left column plots time development of visual patterns as represented in low dimensions using PCA, and the right displays corresponding snapshot sequences at their original pixel level.}
\end{figure}

The first and the second rows in Figure 9 show the time development of the target pattern from the \(0^{th}\) to the \(50^{th}\) steps, and the corresponding closed-loop output generation in the same period, respectively. The third row shows the generation from the \(260^{th}\) step to the \(310^{th}\) step. For each row, the left-hand side plots the time developments of these values in reduced dimensions after applying PCA to the original pixel level representation, and the right-hand side plots sequential snapshots at the original pixel level. These plots indicate that in the early steps (from \(0^{th}\) step to \(50^{th}\) step) the network could produce patterns similar to the target. However, as the steps went by, the output pattern became distorted and its apparent resemblance to the target was lost. The implication here is that during early learning stages, even though the network is yet to form stable limit cycle attractors deeply embedding target patterns in network dynamics, regenerating target patterns for limited cycles is possible through transient dynamics. This observation is revisited in  experiment 2b.

\subsubsection{Experiment 1b - Learning complex patterns}
Experiment 1b investigates the capability of the P-MSTRNN model to learn more complex patterns. A new dataset was prepared, consisting of two types of sequence, each generated by the same 5 people. The sequences concatenated the primitive movement patterns described in section3-1, in this case P1-P2-P3-P4-P5 and P5-P4-P3-P2-P1. Each primitive movement was repeated 4 times, then the next was repeated four times, and the next until the sequence ended. All sequences were around 350 steps long. A network identical to that used in the previous experiment was trained with this new dataset. Training was terminated when closed-loop generation of all training sequences was successful. As with the previous experiments, neural activities were collected while the network was generating closed-loop and PCA was applied to reduce the dimensionality of the data. 

Figure 10 plots neural trajectories generated in the closed-loop mode by different layers of the trained network.

\begin{figure}
  \centering
  \includegraphics[width=0.8\textwidth]{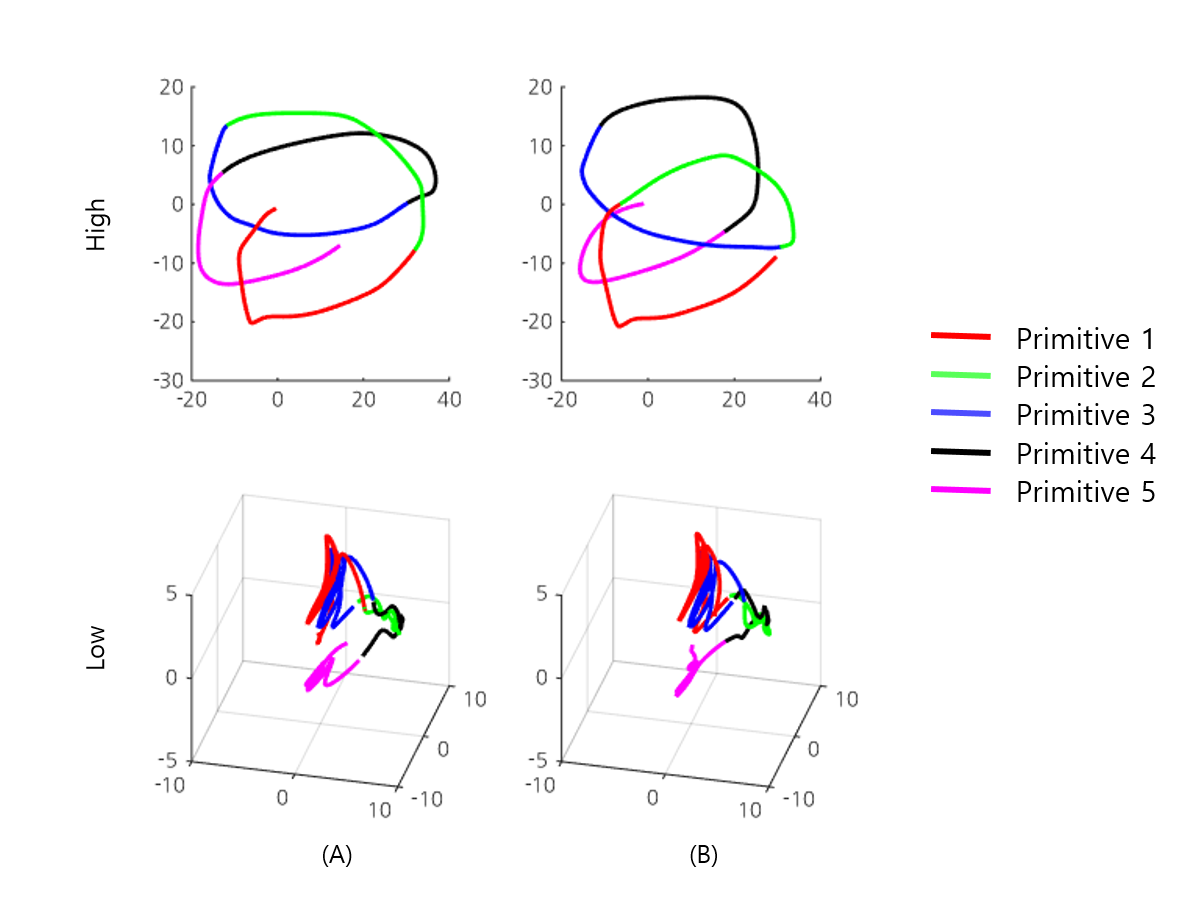}
  \caption{Neural trajectories generated in higher (first row) and lower (second row) levels in the network trained with the concatenated sequence dataset. Dimensional reduction is applied to neural activities recorded. (A) The neural trajectories generating the concatenated sequence P1-P2-P3-P4-P5, and P5-P4-P3-P2-P1 (B). }
\end{figure}

All the data in this figure was recorded when the network performed the closed-loop regeneration of the two target training sequences as performed by subject 1. The first row presents the higher layer’s (\(4^{th}\) layer FM) neural activity and the second row is for the lower layer’s (\(1^{st}\) layer CM) activity. Figure 10 (A) represents activity during closed-loop regeneration of the sequence P1-P2-P3-P4-P5, and (B) the sequence P5-P4-P3-P2-P1. It is worth noting that these neural trajectories were generated as transient ones, beginning from two distinct initial states without cycling or fixation within the periods shown.

In the higher layer, each primitive movement pattern segment (each shown in a different color) appears straighter than those in the lower layer. This is evidence of a cyclic representation of more complex patterns. This result, analogous to those obtained previously (see Figure 7), implies that higher layer activity represents more abstract information (such as the orders of learned primitive movement pattern sequences) while the lower layer activity provides more detailed information.

Interestingly, as shown in Figure 10 (A) and (B), neural trajectories in both lower and higher layers resemble each other, even though their movement pattern sequences are oppositely ordered. This observation points to two important features of present model dynamics. First, the network is capable of segmenting movement patterns built from concatenated training patterns. Second, the network is capable of using hierarchical functions to combine these segmented movement patterns into different concatenations . Third, the network can learn to generate such concatenated sequence patterns using transient neural trajectories.

\subsection{Experiment 2 - Predictive imitation} 
This experiment examines the ability of the P-MSTRNN to predictively imitate various target patterns under various conditions. As described previously, predictive imitation can be performed using two different schemes. On the first scheme, predictive imitation is accomplished using error regression. The activity internal to each layer is modulated in the direction of minimizing error between prediction outputs and target visual input, closed loop (see Figure 2). On the second scheme, the idea of entrainment by inputs can be applied, and the model can be entrained by input. On this scheme, prediction of next step visual inputs is generated from current visual inputs as they are fed into the network, open-loop (see Figure 1.A). It is expected that the model network and the external target system can achieve synchrony by means of phase locking, provided that these two coupled systems share oscillation components. 

Using the data saved from the previous experiment investigating the development of dynamics involved in learning multiple primitive movement patterns generated (differently) by multiple subjects, this experiment clarifies how predictive imitation performance varies depending on different dynamic structures developed at different stages of learning, and whether using the error regression or not. Performance differences between cases trained on the multiple-subject dataset and those trained on the single-subject dataset are investigated for the purpose of examining how variance in exemplar patterns can contribute to generalization in learning, In this case, we tested predictive imitation performance with target patterns demonstrated by previously unknown subjects.  Finally, predictive imitation performance using the error regression is compared with that using entrainment by inputs. 

\subsubsection{Experiment 2a - Predictive imitation with error regression during the developmental process}
Predictive imitation by error regression performance during six different stages of training was assessed and compared using the same network described in experiment 1a. In that experiment, parameters including weights and biases were saved every 100 epochs during 8000 epochs of training. Parameters obtained after training for \(100^{th}\), \(500^{th}\), \(1000^{th}\), \(2000^{th}\), \(4000^{th}\) and \(8000^{th}\) epochs were used in the current experiment. With these six different network states representing stages of learning, we conducted tests on predictive imitation using a test primitive movement pattern sequence generated by three subjects who did not participate in the previous experiments. The test sequence P1-P2-P5-P3-P4-P6-P4-P3-P5-P2-P1 involves multiple switches between the primitive movement patterns described in section 3.1, and was around 2000 video frame/time steps long. Note that this sequence differs from those sequences employed in experiment 1b. We assessed predictive imitation of the target sequence containing unexpected switching between primitive movement patterns by observing the prediction error. This is the error between the network prediction at \(t+2\) and the actual value of input at that time step. The error regression adaptation rate regulating internal state changes was set to 0.1, the window size was set to 20 steps and initial state optimization performed 100 times for every video frame step. 

\begin{figure}
  \centering
  \includegraphics[width=0.99\textwidth]{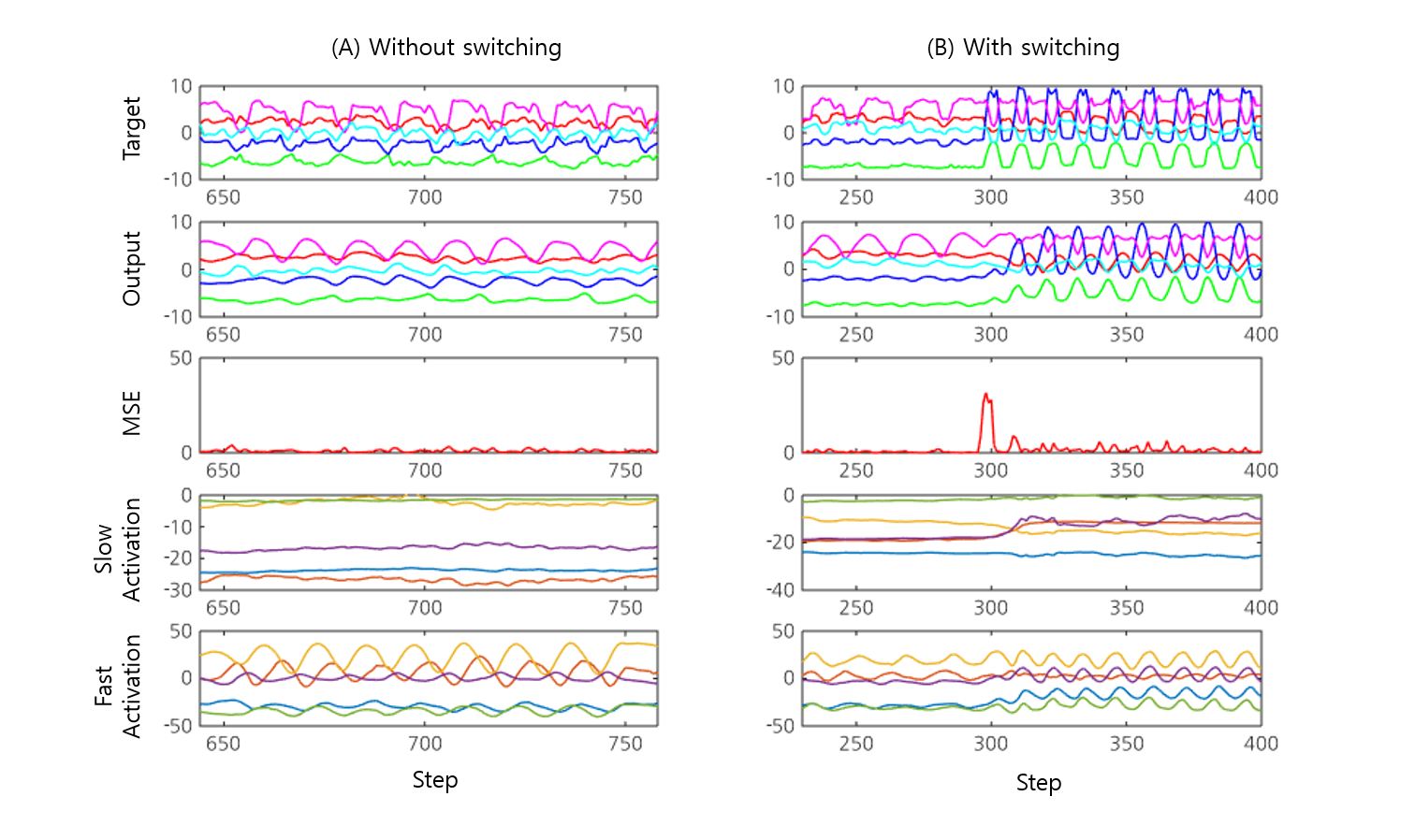}
  \caption{Results of predictive imitation. (A) when the test target sequence consists of only one pattern (P6). (B)  when the primitive movement pattern in the target sequence switches from P2 to P5. The following are shown from the top row down: target visual input, results of the one-step prediction output, MSE, slow activation, fast activation. Fast activation is collected from CMs in the first layer, and slow activation from FMs in fourth layer. All plots (except MSE) present values obtained by PCA. }
\end{figure}

Figure 11 shows a typical example of predictive imitation by error regression \footnote{Supplementary Video2, Video3 at\\ \url{sites.google.com/site/academicpapersubmissions/p-mstrnn }}. This figure shows plots of neural activities after 2000 training epochs (A) when the test pattern included no movement pattern switching, and (B) when the test pattern involved switching from primitive 2 to primitive 5 at around the \(300^{th}\) video frame step. From the top row down, the figure shows the target visual pattern, one-step prediction outputs, mean square prediction error (MSE), and internal activity of both the higher and lower layers. Both the visual inputs and prediction outputs are shown in 5-dimensional values obtained by PCA. The plots for the lower and higher layers show the activities of \(1^{st}\) layer (fast) and \(4^{th}\) layer (slow) FMs after PCA. In Figure 11 (B), MSE rose sharply at around the \(300^{th}\) step, leading to rapid changes in activities in both the lower and the higher layers and a change in the prediction output patterns from P2 to P5. Immediately after these prediction output patterns switch, MSE decreases to near zero. In Figure 11 (A), there is no movement pattern switching in the target sequence, and the error remains low for most of the period. 

The mean square error (MSE) during one-step prediction of the aforementioned target sequences for each stage of learning is given in Table 5.
\begin{table}[]
\centering
\caption{MSE of predictive imitation by error regression. Values presented are averages of three networks trained with different initial random seeds. }
\label{my-label}
\begin{tabular}{ccccccc}
\hline
\begin{tabular}[c]{@{}c@{}}Training  Epoch\end{tabular}                          & 100    & 500    & 1000   & 2000   & 4000   & 8000  \\ \hline
\begin{tabular}[c]{@{}c@{}}Error regression\\ MSE\end{tabular}                    & 0.0468 & 0.0346 & 0.0331 & 0.0320 & 0.0333 & 0.035 \\
\begin{tabular}[c]{@{}c@{}}Limit-cycle Attractors\\ embedding target\end{tabular} & 0      & 0      & 4.5    & 13.75  & 13.5   & 15    \\ \hline
\end{tabular}
\end{table}
The number of limit cycle attractors embedding target patterns for each stage of learning is taken from Table 4. The network trained for 100 epochs could not predict target sequences well, as evidenced by the relatively large MSE. Generated patterns were largely distorted. However, the network trained for 500 epochs could perform one-step prediction quite well, with the additional training decreasing the MSE to 0.0346. All other networks trained for more than 500 epochs could also predict well, with similarly small MSE values. 

In order to account for why the network trained for only 500 epochs (without forming limit cycle attractors deeply embedding the training patterns in network dynamics) can perform predictive imitation with such a relatively small MSE, we monitored look-ahead prediction by error regression in the error regression window during different stages of learning. Figure 12 shows a snapshot of the middle of the on-line error-regression process with a focus on the current “now” step as demonstrated in two networks during two different stages of learning, namely after 500 epochs and after 8000 epochs.

\begin{figure}
  \centering
  \includegraphics[width=0.6\textwidth]{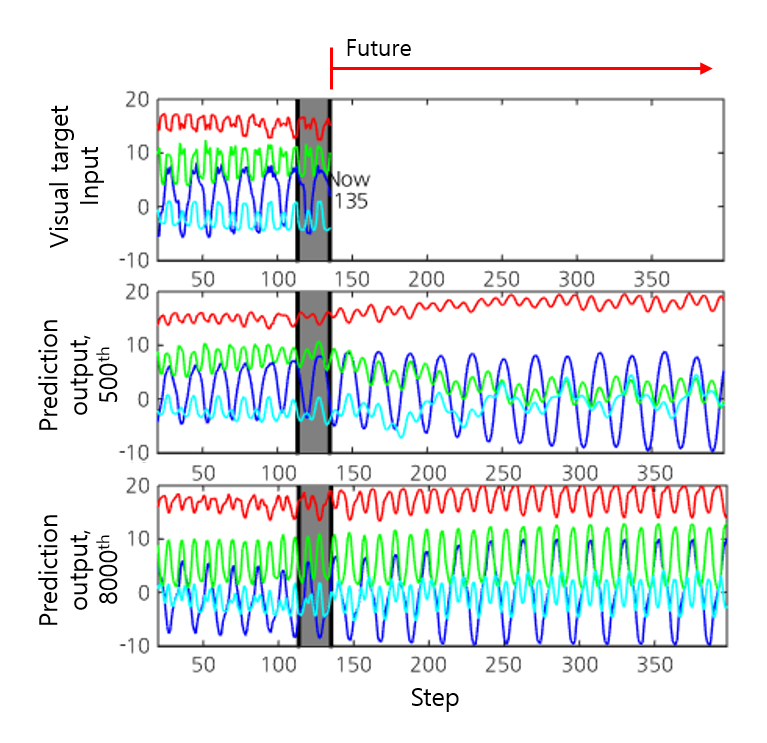}
  \caption{Representative snapshot during error regression. The top row is the target visual input, the second row is the prediction output generated by the network trained for 500 epochs, and the third row is the prediction output generated by the network trained for 8000 epochs. All plots were drawn by PCA applied to the original pixel level representation.}
\end{figure}

These two networks were compared to examine how look-ahead prediction changes depending on network learning stage. The top row in Figure 12 is the target visual input after PCA applied to the pixel level representation. The visual input from the \(0^{th}\) step to the \(135^{th}\) step is of the same category of primitive movement pattern, there is no switching. The second row is the visual prediction output (after PCA) generated by the network trained for 500 epochs, and the third row is the visual prediction output (after PCA) by the network trained for 8000 epochs. The black boxes shown in all plots indicate the error regression (ER) window in the immediate past. In this figure, the current step indicated by “now” is the 135th step, and the ER window 20 steps into the past opens from the \(116^{th}\) step. Here, let us look at how the prediction of future steps, as well as the reconstruction of past steps in the ER window, can be computed by the ER scheme.


The network first performs closed-loop generation starting from the onset of the ER window in the immediate past, from the \(116^{th}\) step to \(135^{th}\) step. Once the closed-loop sequence inside the window is generated, reconstructing the image, this reconstructed image is compared to the target value (the recorded visual input sequence) and this yields error. The error is back-propagated through time steps within the window, eventually adjusting the latent states of the initial step of the window, at the \(116^{th}\) step. Beginning from this step, new closed-loop output is generated, new error computed and back-propagated. This forward generation of closed-loop values and the back-propagation of error within the temporal window, updating the latent states at the onset of the window, is repeated a predefined number of times in order to minimize error between the recorded visual input sequence and closed-loop generation inside the window. Ideally, the network regenerates the target pattern in the ER window with the prediction output sequence when the error is minimized. After the latent state is adjusted by means of this error regression process, look-ahead prediction of visual input is generated for future steps (steps after the \(135^{th}\) step) by means of closed-loop generation. 

Figure 12 shows that dynamics generating look-ahead prediction sequences differ during different training stages. In the network trained for 500 epochs, the prediction output pattern changed gradually from the target pattern as time steps went by, whereas the prediction output pattern in the 8000 epoch training case stayed similar to the target. The prediction output pattern at 500 epochs remains similar to that of the target for at least 10 steps after the “now” step. As stable attractors had not yet developed, this implies that forward closed-loop computation can successfully regenerate the trained cyclic pattern from the onset step of the ER-window to the \(10^{th}\) step by way of transient neural trajectories. This interpretation is supported by the results of experiment 1a, which showed that the network trained for 500 epochs can regenerate trained cyclic patterns by using the transient region for around 50 steps. Furthermore, it is likely that even the network trained for 8000 epochs can successfully regenerate the target pattern using transient neural trajectories because convergence to the limit cycle attractor can require several hundred more steps, as shown in experiment 1a.

In summary, the preceding research found that multiple cyclic movement patterns were retained in transient neural trajectories during early learning phases without forming limit cycle attractors. During later learning stages, target patterns were retained in transient neural trajectories, and limit cycle attractors embedding those patterns formed. It was also found that the transient regions were used in predictive imitation of target patterns, both with networks from earlier and later learning stages. This result is related to \citet{Ahmadi2017} which shows that patterns deviating from exemplar patterns can be learned in transient neural trajectories that converge to attractors embedding the main patterns.

\subsubsection{Experiment 2b - Comparison of multiple subject training and single subject training cases}
Predictive imitation by error regression performance was compared between networks trained in two different conditions, namely that using the dataset generated by multiple subjects (as in the experiments 1a and 2a) and that trained on data from a single subject. We compared performance during all 6 different training stages using the same test target data used in experiment 2a. Table 6 sets out predictive imitation MSE for networks trained under these two conditions during each stage of learning.

\begin{table}[]
\centering
\caption{MSE of predictive imitation by error regression under different training conditions}
\label{my-label}
\begin{tabular}{ccccccc}
\hline
\begin{tabular}[c]{@{}c@{}}Training  Epoch\end{tabular}              & 100    & 500    & 1000   & 2000   & 4000   & 8000   \\ \hline
\begin{tabular}[c]{@{}c@{}}Multi-subject training\\ MSE\end{tabular}  & 0.0468 & 0.0346 & 0.0331 & 0.0320 & 0.0333 & 0.035  \\
\begin{tabular}[c]{@{}c@{}}Single-subject training\\ MSE\end{tabular} & 0.0476 & 0.0428 & 0.0425 & 0.0411 & 0.0405 & 0.0364 \\ \hline
\end{tabular}
\end{table}

The MSE of the multiple subjects case is smaller than that of the single subject case during every stage of training. This might be because larger variation in training patterns enhances generalization.

\subsubsection{Experiment 2c - Predictive imitation via entrainment by input}
This final experiment compares two possible schemes for predictive imitation, one by error regression and the other input entrainment. The same network used in experiment 1a, informed by the same network parameters (such as connectivity weights and biases) recorded during different stages of training in experiment 1a, was compared with the same network informed by the test data used in experiment 2a at each corresponding stage of development.

Predictive imitation by input entrainment uses the open-loop operations described above (see Figure 1 (A)). In the case of a sudden switch in movement pattern, internal dynamics resituate to the current context by means of entrainment to current visual input \citep{Kelso1997, Taga1991}. Figure 13 (A) and (B) show typical predictive imitation results using the entrainment by inputs mechanism at 2000 epochs. 

\begin{figure}
  \centering
  \includegraphics[width=0.99\textwidth]{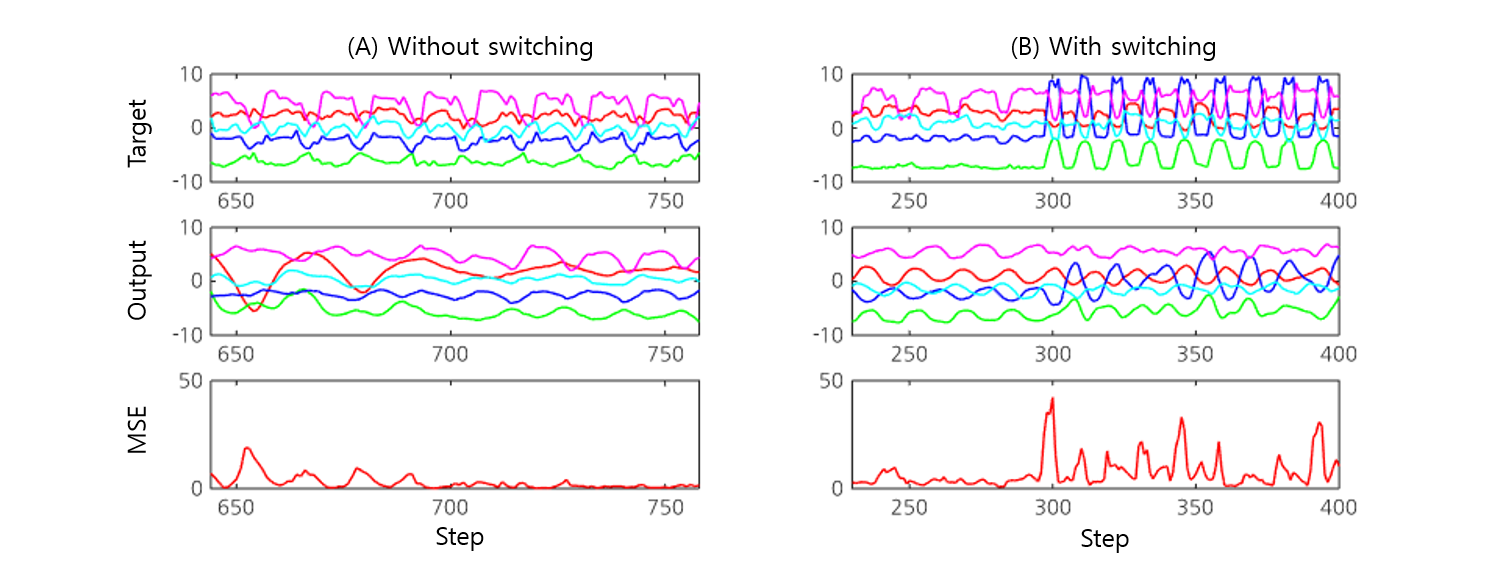}
  \caption{Examples of predictive imitation by entrainment by input. (A) when the test sequence contains no switching (only P6), (B) when the test sequence contains switching from P2 to P5. Target visual input (first row), corresponding prediction output (second row) and MSE (third row) are plotted. The prediction outputs and the target visual inputs are plotted by PCA of original pixel level representation. }
\end{figure}

Figure 13 (A) and (B) shows the cases with and without switching of movement patterns, respectively. As with previous figures, both the targets and the prediction outputs are plotted by PCA applied to the pixel level representation. Compared to the performance of the scheme using error regression as shown in Figure 11 (A), prediction outputs shown in Figure 13 (A) are distorted and generate larger error. Clearly, using error regression to update anticipated situations, in effect priming the network for probable appropriate action, outperforms entrainment by input in cases without movement pattern switching. With switching, entrainment by input performs even worse as shown in Figure 13 (B). Output prediction generated significant error during and after the visual target input switched from P2 to P5 at around the 300th step. The degree of distortion in output predictions was significantly larger than that of the error regression scheme (see Figure 11 (B).) 

A quantitative analysis supporting this conclusion is set out in Table 7.

\begin{table}[]
\centering
\caption{MSE of ER and entrainment by input at different training stages}
\label{my-label}
\begin{tabular}{ccccccc}
\hline
\begin{tabular}[c]{@{}c@{}}Training  Epoch\end{tabular}  & 100    & 500    & 1000   & 2000   & 4000   & 8000   \\ \hline
\begin{tabular}[c]{@{}c@{}}ER  MSE\end{tabular}          & 0.0468 & 0.0346 & 0.0331 & 0.0320 & 0.0333 & 0.035  \\
\begin{tabular}[c]{@{}c@{}}Entrainment  MSE\end{tabular} & 0.0625 & 0.0457 & 0.0452 & 0.0515 & 0.0620 & 0.0680 \\ \hline
\end{tabular}
\end{table}

Table 7 compares the MSE generated by the error regression scheme and the entrainment by inputs scheme. MSE generated by the entrainment by inputs scheme was significantly larger than that of the error regression scheme. Predictive imitation using error regression outperforms the entrainment by inputs.

\section{Discussion and Conclusion}
In the present study, we proposed a novel predictive coding model, the predictive spatio-temporal scales recurrent neural network (P-MSTRNN). We observed the development of its internal dynamics during training on a dynamic visual stream dataset depicting whole body cyclic primitive movement patterns as demonstrated by multiple subjects. We assessed predictive imitation performance under various conditions, using multiple and single subject datasets, and using entrainment by input and predictive imitation by error correction.

Analysis of the development  of network dynamics showed that the number of attractors embedding target patterns increased as learning proceeded. The higher layer developed fixed point attractors, and the lower developed limit cycle attractors. Analysis of network dynamics at the end of the whole learning period showed that a functional hierarchy had been developed in the network, with the upper and lower layers working in tandem to predictively imitate target patterns. Each of the fixed points generated in the higher layer represented the intention for a specific primitive movement pattern, and played the role of bifurcation parameter by inducing a transition in the lower layer network to a corresponding limit cycle attractor. Interestingly, a similar functional hierarchy was observed even in earlier stages of learning, in these cases exploiting the transient regions of the developed dynamic structure instead of the invariant set of limit cycle attractors. Also interesting was the observation that subject-wise idiosyncrasies in primitive movement patterns were generated for a few cycles by way of transient trajectories during early learning. Finally, when the network was trained with different sequences of primitive movement patterns, neural trajectories for the same movement pattern exhibited similar trajectories regardless of sequence order. This suggests that the network was able to compose/decompose sequences with/into their primitive segments using the functional hierarchy developed through learning. 

The second part of the present study assessed predictive imitation performance during different learning stages. Using the error regression scheme, performance in terms of prediction error in the earlier stage (500 epochs, without forming limit cycle attractors embedding target patterns) was similar to performance at the later stage (8000 epochs, with fully formed limit cycle attractors embedding target patterns). Furthermore,  the network trained with the multiple subject dataset outperformed that trained with the single subject dataset during predictive imitation. This suggests that generalization in learning is achieved by training with diverse patterns, and that this generalized learning facilitates predictive imitation by strengthening inferences of target patterns. Lastly, two predictive imitation schemes were compared, namely that by error regression and by input entrainment, with performance of the former far superior to that of the latter.

\subsection{Related studies}
The present study builds on various ideas inherited from previous studies. The idea of using the error back-propagation through time (BPTT) algorithm \citep{Rumelhart1985} to drive active inference using on-line error regression came from \citet{Tani1999, Tani2003}. \citet{Tani1999} proposed a hierarchically organized predictive coding type model with localized representations, by using a mixture of gated RNNs at each level of the network. Simulated mobile robots using this model infer the gate opening at each level by means of the error regression and can recognize the ongoing perceptual flow as segmented in multiple levels by this mechanism. Tani and colleagues \citep{Tani2003, Ahmadi2017} have demonstrated similar active recognition function using a distributed representation of contextual activities in a hierarchically-organized RNN model which infers these contextual activities in multiple levels by way of on-line error regression using BPTT.

One point of interest is that these models use the same connectivity pathways but in opposite directions for top-down forward activation and bottom-up backward error propagation. Other predictive coding models such as those proposed by \citet{Rao1999} and by \citet{Friston2005} use different pathways for forward activation and backward error propagation. Future study should explore differences between these two schemes both qualitatively and quantitatively.

Recently, predictive dynamic vision models have been proposed \citep{Lotter2016, Srivastava2015, Ranzato2014, Finn2016}. Although these models predict future pixel frames in visual sequences by learning predictive models from given datasets, they cannot perform active inference using the prediction error signal as the P-MSTRNN does. One exception could be PredNet proposed by Lotter and colleagues \citep{Lotter2016}. This model uses different pathways to propagate  top-down prediction and bottom-up error signal after prediction. The PredNet used a natural color video dataset for prediction experiments and the predictions were conducted for short frame ranges of fewer than 20 frame steps, and does not show how prediction error can be used for active inference of given target visual streams.

\subsection{Can transient dynamics be used for memory dynamics?}
General human movements are repetitive, consisting of frequently used movement primitives such as reaching, grasping, walking, waving and others. Some researchers (for example \citep{Ijspeert2013}) suggest that movement primitives are either discrete movements such as reaching or cyclic movements such as walking or shaking, considered to be developed in terms of fixed point dynamics and limit-cycle dynamics, respectively. In the current paper, we focused on movement patterns based on limit-cycle dynamics. 
One interesting finding in the current study is that, during learning, transient dynamics can embed target cyclic movement patterns. Analysis of the dynamical structures developed at different stages in the learning processes showed that all of the subject-wise movement patterns could be regenerated during early video frame steps by each initial state taken from each stage of the training process, although many of these patterns decayed as steps went by because they were not embedded in stable limit cycle attractors. 

Analogous observations have been made in previous studies \citep{Ahmadi2017, Nishimoto2004}. Ahmadi and Tani reported that a MTRNN model which had been trained with a set of prototypical cyclic patterns of varying  amplitude and periodicity can generate as well as recognize similar prototypical patterns, and incorporate diverse joint variance in amplitude and periodicity by using the transient regions of developed dynamics. At the same time, they reported that the network trained with only those prototypical patterns, without any variations, could not deal with fluctuations in such patterns. Nishimoto and Tani (2004) presents similar results in a simulation experiment in which a deterministic RNN model was trained with probabilistic symbol sequences generated by a particular finite state machine (FSM). It was shown that such probabilistic sequence patterns could be regenerated by the trained network using transient chaos. However, the trajectories generated by means of transient chaos tended to converge into limit cycle attractors, at which point the network can no longer generate target patterns produced by the FSM. 

Along with these prior studies, the current study also suggests that transient dynamics can be used for memorizing and recalling dynamic patterns. This is interesting because exploiting transient dynamics runs contrary to conventional dynamical systems thought, which emphasizes attractor formation for memorizing dynamic patterns \citep{Kelso1997, Beer1995}. But, using transient dynamics for memorizing dynamic patterns should prove a useful approach in some artificial systems applications because necessary training periods can be drastically reduced. Indeed, the current study shows that it is not necessary to wait 8000 training epochs until target patterns are embedded as limit cycle attractors in order to successfully perform predictive imitation as well as the closed-loop generation of primitive movement patterns. Instead, training for just 500 epochs is enough to perform those tasks.

\section{Future work}
The current study used training datasets of limited complexity, with body movement patterns composed according to artificial rules. Future research should overcome these limitations. 
Also, video images were limited to only 36x36 binary. This limitation is partly due to computational requirements. Predictive coding at the pixel level incurs significant computational costs for learning even in the current low resolution, as compared to learning categorization without reconstruction of the image. The current training of six patterns with five subjects costs almost two weeks computing time. We assume that the novelty of the current paper is that it may be the first paper to deal with learning multiple human movement visual patterns from multiple subject data. Therefore, we put more efforts on analysis of the model dynamics by using relatively simple learning targets. Binary image video facilitated reliable analysis. More complex cases can be pursued now after understanding the underlying mechanism. Future study should investigate the capabilities of the model after training with more natural human action patterns, for example grayscale performance videos such as those publically available on Youtube.
It is also important to investigate how the current model based on deterministic dynamics can deal with probabilistic properties latent in natural image datasets. From the view point of Bayesian modeling, statistical structures underlying sequential data can be extracted as a probability distribution function of random variables through learning. On the other hand from the deterministic dynamics view, one may consider that such statistical structures can be learned and embedded in deterministic chaos \citep{Tani1994, Namikawa2011}.


\subsection*{Acknowledgments}
This work was supported by a National Research Foundation of Korea (NRF) grant funded by the Korea government (MSIP) (No.2014R1A2A2A01005491).

\end{document}